\documentclass[lettersize,journal]{IEEEtran}
\IEEEoverridecommandlockouts
\usepackage{cite}
\usepackage{hyperref}
\usepackage{multirow}
\usepackage{amsmath,amssymb,amsfonts}
\usepackage{graphicx}
\usepackage{textcomp}
\usepackage{xcolor}
\usepackage{comment}
\usepackage{graphicx}
\usepackage{array}
\usepackage{xcolor}
\usepackage{tabularx}
\usepackage{booktabs}
\usepackage{makecell}
\usepackage{mathtools}
\usepackage{rotating}

\usepackage{amsmath,amsfonts}
\usepackage{array}
\usepackage[caption=false,font=normalsize,labelfont=sf,textfont=sf]{subfig}
\usepackage{textcomp}
\usepackage{stfloats}
\usepackage{url}
\usepackage{verbatim}
\usepackage{graphicx}

\usepackage{algorithmicx}
\usepackage{algorithm}
\usepackage{algpseudocode}

\def\BibTeX{{\rm B\kern-.05em{\sc i\kern-.025em b}\kern-.08em
    T\kern-.1667em\lower.7ex\hbox{E}\kern-.125emX}}
\begin{document}


\title{LiteEvent-AE: Lightweight Autoencoder for Event-Based Vision on Low-Latency Energy-Constrained Edge Devices}


\author{
Riadul~Islam,~\IEEEmembership{Senior Member,~IEEE},
Joey~Mulé, 
Dhandeep Challagundla,~\IEEEmembership{Student Member,~IEEE},
Shahmir Rizvi, 
Sean Carson,
and Rachit~Saini

\thanks{R Islam, J. Mulé, D. Challagundla, S. Rizvi, S. Carson and R. Saini are with the Department 
of Computer Science and Electrical Engineering, University of Maryland, Baltimore County, 
MD 21250, USA e-mail: {riaduli@umbc.edu}.}

\thanks{This work was supported in part by the National Science Foundation (NSF) award number: 2138253, the Maryland Industrial Partnerships (MIPS) program under award number MIPS0012, and the UMBC Startup grant.}
\thanks{Copyright (c) 2025 IEEE. Personal use of this material is permitted. 
However, permission to use this material for any other purposes must be 
obtained from the IEEE by sending an email to pubs-permissions@ieee.org.}
}
\markboth{IEEE Canadian Journal of Electrical and Computer Engineering Arxiv} 
{Shell \MakeLowercase{\textit{et al.}}: ??????}

\newcommand{\fixme}[1]{{\Large FIXME:} {\bf #1}}

\maketitle

\begin{abstract}
Event-based vision has emerged as a promising paradigm for energy-aware artificial intelligence (AI), offering sparse, low-latency visual signals that reduce redundant data processing and support sustainable edge computing. However, the asynchronous and noise-prone nature of event streams creates challenges for conventional deep learning models, which are often too computationally intensive for low-power embedded platforms. This work presents a compact and configurable event-driven autoencoder that efficiently compresses neuromorphic data while preserving essential spatiotemporal structure for downstream inference. The architecture integrates lightweight convolutional encoding with robust performance under adaptive event thresholding and a minimal classifier head, enabling substantial reductions in computational cost without degrading recognition fidelity.
Extensive evaluations on the Smart Event Face Dataset (SEFD) and Event-Based Crossing Dataset (EBCD) show that the proposed framework achieves competitive or superior accuracy compared to  YOLOv9 while requiring up to 35.6$\times$ fewer parameters. To assess real-world sustainability, the model is deployed on resource-constrained hardware: a Raspberry Pi 4B and a NVIDIA Jetson Nano. On NVIDIA Jetson Nano, it delivers real-time throughput of 44.8 FPS. On a Raspberry Pi 4B CPU, the 50\% autoencoder classifier consumes 16.19 J for the evaluated inference workload, corresponding to approximately 726.3$\times$ lower energy consumption than YOLOv9 under the same evaluation protocol. These results demonstrate the potential of compact event-driven models to advance environmentally conscious, low-power AI systems for high-speed perception in autonomous, mobile, and embedded computing environments.

\end{abstract}


\section{Introduction}
\label{sec:intro}

\begin{figure}[t]
		\begin{center}
			\includegraphics[width = 0.49\textwidth]{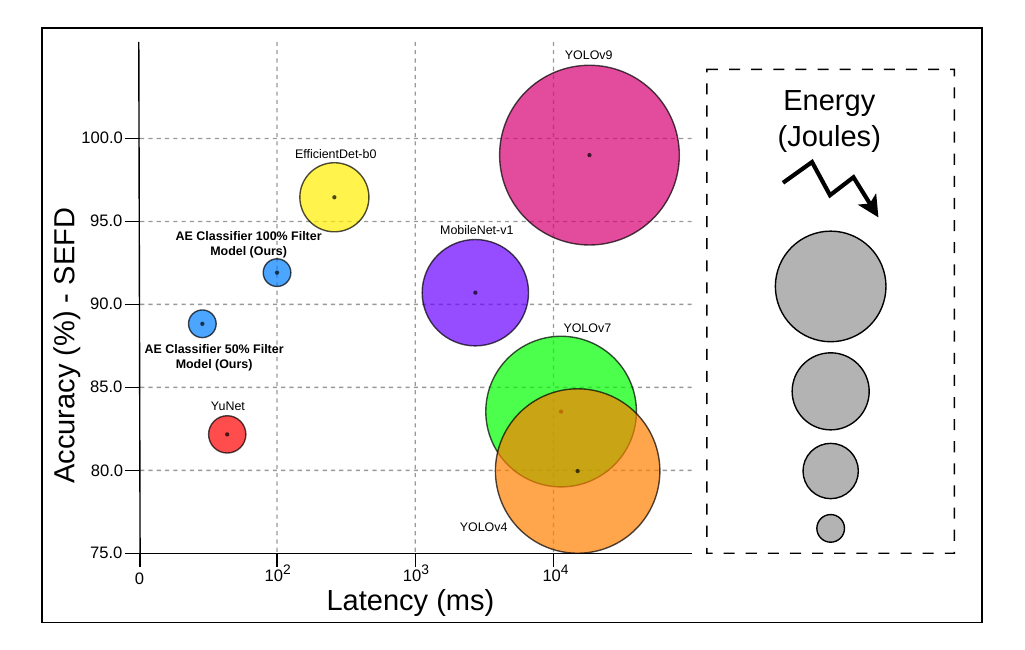}
			\vspace{-0.6cm}
			\caption {Comparison of average model accuracy, latency, and energy consumption on the SEFD~\cite{Islam_descriptor:2024} dataset, evaluated on a Raspberry Pi 4B CPU.}
            \label{fig:motivation}
			\vspace{-0.75cm}
		\end{center}
\end{figure}

Rapid visual perception is increasingly essential in emerging technologies that rely on instantaneous scene understanding and decision-making, such as 
autonomous vehicles, robotic systems, industrial monitoring, and scientific experimentation. Conventional frame-based imaging architectures, however, are fundamentally limited in high-speed scenarios, often producing blurred images, delayed responses, and redundant visual data when confronted with fast or complex motion. To overcome these bottlenecks, event-driven vision sensors have been developed to capture dynamic visual information in a more biologically inspired manner. Instead of recording full image frames at fixed time intervals, these sensors asynchronously register per-pixel brightness changes, enabling continuous tracking of motion with exceptionally low latency and high temporal fidelity~\cite{PROPHESEE:2024, Joey_ebcd:2025}.

Despite their advantages, the unconventional output of event-driven sensors poses unique challenges for modern artificial intelligence frameworks. Unlike dense, regularly sampled image frames, event streams are sparse, asynchronous, and contain varying levels of noise, making direct application of conventional convolutional neural networks (CNNs) inefficient and often inaccurate. Moreover, real-time deployment in embedded or edge devices requires models that are both computationally lightweight and power-efficient without compromising detection precision~\cite{Islam_icrest:2023}. 
These constraints motivate the development of specialized learning architectures capable of effectively encoding the spatiotemporal structure of event data. In this context, we introduce a lightweight event-based autoencoder that encodes and reconstructs asynchronous spike streams while  preserving 
the essential motion dynamics embedded in the data and structural cues, thereby enabling robust, low-latency object classification in high-speed, 
resource-constrained environments. 
Figure~\ref{fig:motivation} compares the accuracy, latency, and energy efficiency of state-of-the-art models with the proposed event-based autoencoder (AE) classifier architectures on the Smart Event Face Dataset (SEFD)~\cite{Islam_descriptor:2024}. 



Autoencoders' unsupervised learning capability further enables adaptation to diverse event-driven scenarios, making them valuable for low-power, high-speed vision applications such as robotics, autonomous navigation, and industrial automation~\cite{Wang_auto:2016, Tai_pottsmgnet:2024, gruel2022event}.

Furthermore, recent work has explored spatiotemporal transformers for streaming object detection~\cite{li2023sodformer}. Although effective on optical frame and flux-based representations, these approaches introduce substantial computational overhead and typically achieve limited mean average precision (mAP). Other efforts have investigated binary event history images (BEHIs) 
combined with lightweight CNN encoders for high-speed object detection~\cite{wang2022ev}; however, these solutions incorporate auxiliary modules for uncertainty quantification
to resolve motion-induced ambiguities. Despite these advances, developing 
scalable architectures that natively adapt to the asynchronous, data-driven nature of event vision remains an open research challenge.

Although spiking neural networks (SNNs)~\cite{islam2024benchmarking, kamata2022fully} are commonly used in event-based vision due to 
their time-dependent processing, their temporal dynamics introduce 
considerable complexity. Our approach instead leverages static event frames that can be processed with standard convolutions, introducing a static event autoencoder–based classification architecture to address these limitations.
The proposed architecture jointly addresses event representation efficiency, latent feature compression, and edge inference complexity.
The architecture specializes in event-based detections, enabling an adaptive threshold selection mechanism for energy-efficient computer vision tasks with event cameras. Hence, event sensors can adaptively select thresholds to dynamically adjust event detection sensitivity based on environmental conditions and signal characteristics, effectively reducing redundant event generation while preserving essential visual information. In particular, the main contributions of this research are:
\begin{itemize} \renewcommand{\labelitemi}{$\bullet$}
	   \item We introduce a configurable and lightweight autoencoder~\cite{islam2025eaeventautoencoderhighspeed} architecture tailored for event-based object classification.
       \item We demonstrate the scalability and robustness of the architecture across multiple event activity thresholds, validating its adaptability to varying noise levels and scene dynamics.
          \item We evaluate the hardware efficiency of the proposed models on embedded platforms such as Raspberry Pi 4B and NVIDIA Jetson Nano, showcasing real-time performance under tight resource constraints.
	   \item We compare with SOTA CNNs, such as YOLOv9~\cite{wang2024yolov9}. Our proposed event classifier achieves up to 35.6$\times$ fewer parameters, 22.07× higher frame rates (FPS), and over 700× lower energy consumption.
    \end{itemize}


\section{Background}
\label{sec:background}


\subsection{Existing Event Vision Applications}
\label{subsec:autoencoder_application}
Researchers have applied event vision to many different tasks, including: automotive~\cite{prophesee_gen1:2020, prophesee_1Mpx:2020, Joey_ebcd:2025}, surveillance~\cite{Miao_neuromorphic:2019}, object classification~\cite{Orchard_NCaltec:2015}, gesture recognition~\cite{Amir_gesture:2017, Bi_graph:2019}, action detection~\cite{Reddy_recognizing:2013}, flow detection~\cite{Zhu_flow:2018}, and face detection~\cite{Islam_descriptor:2024}. With these efforts, both industry and academia generate large-scale event datasets, which can be categorized into event-sensor-based datasets~\cite{prophesee_gen1:2020, prophesee_1Mpx:2020, Binas_ddd17:2017} and simulation-based event-representation datasets from conventional frame-based sensor data~\cite{Joey_ebcd:2025, Islam_descriptor:2024, Reddy_recognizing:2013}. However, most event vision datasets concentrate on single-threshold data. In this research, we concentrate on multi-threshold datasets~\cite{Joey_ebcd:2025, Islam_descriptor:2024} to characterize a robust autoencoder model. Adaptive threshold selection dynamically modulates event detection sensitivity in response to varying environmental conditions and signal characteristics. This approach substantially reduces unnecessary event emissions without compromising the essential visual content needed for downstream processing. By optimizing the event representation pipeline, it enhances the computational efficiency and energy performance of event-based vision systems, thereby improving their suitability for power-constrained applications such as robotics, autonomous vehicles, and wearable technologies.

\subsection{Existing Event Vision Object Classification Methods}
\label{subsec:existing_methods}
Event vision object detection models are characterized by dense or sparse event representations~\cite{ZhuVoxel,RebecqVoxelFrame}. Sparse representations seek to exploit temporal information in a continuous event stream. Dense event representations coalesce asynchronous event streams into structured, frame-like tensors that can be processed using standard vision techniques. Because these representations resemble conventional image frames, they can often leverage advances from traditional computer vision research~\cite{torbunov2025evrt}. Dense event representations benefit from straightforward hardware support and simpler integration with existing vision architectures. In contrast, sparse representations aim to exploit the fine-grained temporal structure of event streams~\cite{zubic2023chaos}. Current state-of-the-art event-vision detectors commonly adopt transformer or hybrid transformer–CNN backbones, reflecting a trend toward frame-like processing even in event-based pipelines. A fundamental challenge, however, is that event cameras produce no output for static scenes; if both the camera and object remain stationary, the object effectively vanishes. To compensate for this lack of object permanence, some recent models incorporate recurrent modules to retain temporal context~\cite{torbunov2025evrt}.

\subsection{Existing Evaluation Methods}
\label{subsec:existing_evaluation_methods}
Evaluation of event vision object detection models parallels traditional computer vision evaluation methods. SOTA event-vision datasets include the NMNIST~\cite{OrchardNMNSIT2015}, DVSGesture~\cite{AmirDVSGesture2017}, Gen1~\cite{prophesee_gen1:2020}, 1Mpx \cite{prophesee_1Mpx:2020}, SEFD~\cite{Islam_descriptor:2024}, and event-based crossing dataset (EBCD)~\cite{Joey_ebcd:2025}. The performance of the models is typically characterized by MS COCO evaluation metrics~\cite{lin2014microsoft}, parameter count, and runtime.

Recent studies have shown that autoencoder-based deep learning frameworks can effectively address data sparsity challenges in multimodal recommendation scenarios~\cite{rajput2024autoencoder}. CNN-based autoencoders have likewise been employed for tasks such as medical image compression and diagnostic classification~\cite{fettah2024convolutional}.

Several studies have investigated recurrent neural network (RNN)–based autoencoder architectures for optical-flow estimation in event-driven vision systems~\cite{Wan_event_auto:2022, Li_rnn_auto:2020, Gehrig_rnn_auto:2021, Hidalgo_rnn_auto:2020, Yu_spikingvit:2025, li2022asynchronous, kim2022ev, zhang2022spiking}. By exploiting the inherent sequential modeling capabilities of RNNs, these approaches effectively encode temporal dependencies in event streams, making them particularly suitable for motion estimation and tracking applications. Predicting optical flow is a key function within event-driven vision systems, enabling accurate motion analysis in dynamic environments by estimating object motion from sparse, asynchronous event data. Prior studies have employed multi-encoder architectures for event-based fusion~\cite{Han_auto_fusion:2020, Zhang_auto_fusion:2021}. 


Nevertheless, despite their demonstrated capabilities, these methodologies are often ill-suited for low-cost, high-speed embedded platforms. The substantial computational demands and elevated power requirements associated with RNN-based models and encoder-fusion pipelines present considerable obstacles for deployment in resource-limited environments, including edge devices and embedded systems. In contrast to energy-aware computing strategies proposed in prior work~\cite{islam_dcmcs:2018, Kodukula_dvfs:2023, Islam_tcasii:2021, Challagundla_tvlsi:2024, islam2011high}, the adoption of such architectures on low-power event sensors may incur excessive latency and reduced energy efficiency, thereby limiting their feasibility in practical settings where power constraints are paramount.
To overcome these challenges, this study introduces a scalable architectural framework that leverages threshold-based event data to enable adaptive thresholding, thereby supporting energy-efficient visual processing in event-driven imaging systems.

\color{black}{
\section{Proposed Methodology}

\begin{figure*}[t!]
		\begin{center}
			\vspace{-0.7cm}
			\includegraphics[width = 0.75\textwidth]{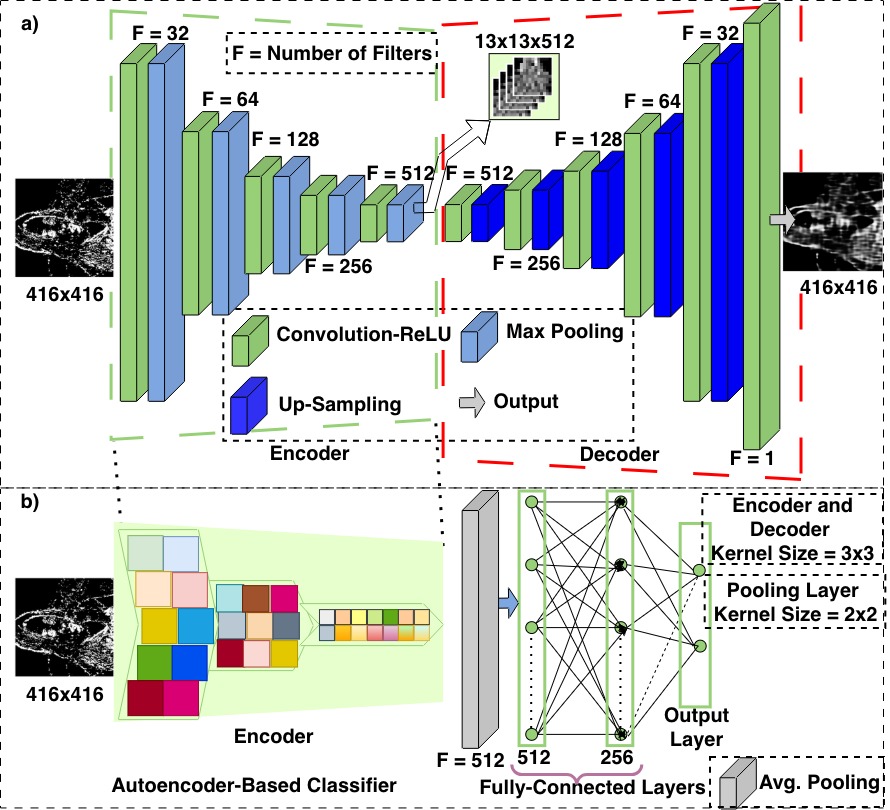}
			\vspace{-0.3cm}
			\caption {(a) {The proposed event autoencoder uses a conventional convolution layer followed by ReLU and Max pooling layer on the encoder side, while the decoder uses a similar convolution layer followed by ReLU but an up-sampling layer; and (b) the autoencoder-based classifier concatenates the encoder with two fully connected layers.}
				\label{fig:autoencoder}}
			\vspace{-0.75cm}
		\end{center}
\end{figure*}
An autoencoder, in the context of computer vision, is a neural network architecture that learns compact, meaningful visual feature representations by reconstructing an input image from a compressed latent space. It uses convolution at its core and consists of two main components: an encoder that transforms high-dimensional image data into a lower-dimensional feature embedding, and a decoder that reconstructs the original image from this embedding. Through this reconstructive learning process, autoencoders implicitly capture essential visual structures such as shape, texture, intensity, and spatial patterns, making them useful for tasks like denoising~\cite{Zhou_denoising:2024}, anomaly detection, dimensionality reduction~\cite{Kim_tied:2024}, and unsupervised feature learning~\cite{Ustek_anomaly:2024}. Their ability to discover latent representations without labeled data makes them a foundational tool in modern computer vision research and applications.

Autoencoders are powerful and flexible neural networks capable of learning highly discriminative features from input data while maintaining a simple and efficient architecture.  By compressing high-dimensional input data into a compact latent representation, autoencoders remove redundant information while preserving critical patterns necessary for downstream tasks~\cite{Chen_auto_cvpr:2023}.

The proposed architecture 
can be interpreted as a system that compresses and transmits visual data through a constrained communication channel. The encoder applies a nonlinear transformation that maps high-dimensional images to a low-entropy latent representation, performing lossy compression that captures salient structures such as edges, textures, and shapes, while discarding redundant pixel-level information. The decoder reconstructs the input from this compressed code, enforcing a trade-off between information bottleneck constraints and reconstruction fidelity. 
For the $l$-th convolutional layer in the encoder, the feature map is computed as
\begin{equation}
 h^{(l)} = \sigma \big( W^{(l)} * h^{(l-1)} + b^{(l)} \big).
\end{equation}
Here, $*$ denotes the convolution operation, $W^{(l)}$ and $b^{(l)}$ are the learnable convolution
kernels and biases, and $\sigma(\cdot)$ is a nonlinear activation function.
The initial feature map is the input image $h^{(0)} = x$. The latent representation produced by the
encoder is $z = h^{(L)}$.

The decoder reconstructs the input by progressively upsampling the latent representation using
transposed convolutions. For the $l$-th decoder layer, the operation is given by
\begin{equation}
    \hat{h}^{(l-1)} = \sigma \big( \tilde{W}^{(l)} * \hat{h}^{(l)} + \tilde{b}^{(l)} \big).
\end{equation}
Thus, $\tilde{W}^{(l)}$ and $\tilde{b}^{(l)}$ are the
decoder's learnable parameters, and $\sigma(\cdot)$ is an activation function.
The final reconstruction is obtained as
$\hat{x} = \hat{h}^{(0)}$.

The proposed event autoencoder is a compact convolutional encoder–decoder designed specifically for dense event representations, as shown in Figure~\ref{fig:autoencoder}(a). 
For real-time streaming pipelines, this approach also requires an event accumulation stage (event-to-frame conversion), which is natively supported by event-sensors and enables compatibility with existing GPU-accelerated CNN frameworks.
The encoder progressively condenses each event frame into a low-dimensional latent representation using a stack of convolutional blocks with Rectified Linear Unit (ReLU) activation and spatial downsampling. 
The ReLU activation function plays a critical role in stabilizing and enhancing feature extraction. First, ReLU introduces nonlinearity while maintaining computational efficiency, allowing neural networks to model the complex spatiotemporal patterns inherent in asynchronous event streams. Because event data are sparse and often exhibit high temporal resolution, ReLU’s simple thresholding (setting negative activations to zero) helps suppress noise, improving robustness to spurious events. Additionally, ReLU promotes sparse activations, which aligns well with the inherently sparse nature of event data and reduces unnecessary computational overhead. This sparsity not only improves training efficiency but also enhances the network’s ability to highlight meaningful changes in the scene, ultimately supporting more discriminative and energy-efficient processing in event-based vision tasks.

The decoder mirrors this process by upsampling and refining the latent features to reconstruct the original event frame. By enforcing a reconstruction objective during training, the network learns to capture salient spatial structure and activity patterns characteristic of event data while discarding high-frequency noise and threshold-induced artifacts. This results in a latent space well aligned with downstream classification tasks.

To convert the autoencoder into a classifier, shown in Figure~\ref{fig:autoencoder}(b), the pretrained encoder is repurposed as a fixed feature extractor, and two fully connected layers are appended to the latent output. This separation cleanly decouples representation learning from task-specific discrimination: the encoder provides a stable event-driven embedding, while the lightweight classifier learns decision boundaries without modifying the trained spatial transform. This structure is particularly effective for resource-constrained platforms, as it minimizes parameter count and enables real-time performance on devices such as Raspberry Pi 4B and Jetson Nano.

\section{Experimental Results and Discussion}

\paragraph{Experimental Setup} For this study, we employed the SEFD~\cite{Islam_descriptor:2024}, a publicly available event-based facial dataset derived from the Aff-Wild collection~\cite{Kollias:2019} for training and testing. The Aff-Wild dataset contains 298 video sequences totaling more than 1.22 million frames, offering extensive variability in facial expressions associated with different emotional states and rapid affective transitions. It further encompasses a wide range of head poses, illumination conditions, and facial occlusions, thereby providing a comprehensive benchmark for robust facial analysis~\cite{Kollias:2019}.
The SEFD dataset contains one class, persons, specifically facial regions. We also used another dense event vision dataset, the event-based crossing dataset (EBCD)~\cite{Joey_ebcd:2025}, which consists of an automotive and pedestrian crossing task derived from the frame-captured, NTU Pedestrian Dataset~\cite{NTU:2019}. The EBCD dataset, however, samples a more extensive range of thresholds. The EBCD dataset contains two classes: pedestrians and vehicles.
The training uses conventional binary cross-entropy (BCE) loss. The average BCE loss for $N$ number of samples can be calculated using $\mathcal{L}{\mathrm{BCE}}=-\frac{1}{N}\sum_{i=1}^{N}\left[y_i\log(\hat{y}_i)+(1-y_i)\log(1-\hat{y}_i)\right]$, where $y_i$ and $\hat{y}_i$ are the actual label and predicted probability, respectively. For training and testing on the proposed and baseline methods, we used standard NN training and testing procedures using the same threshold-generated event frames, derived from source benchmark datasets (i.e., SEFD~\cite{Islam_descriptor:2024} and EBCD~\cite{Joey_ebcd:2025}).
For both the autoencoder and the autoencoder-based classifier, the following hyperparameters were used during training: batch size = 32, number of epochs = 100, and learning rate = 0.001.
\begin{figure}[t!]
\begin{center}
\includegraphics[width = 0.5\textwidth]{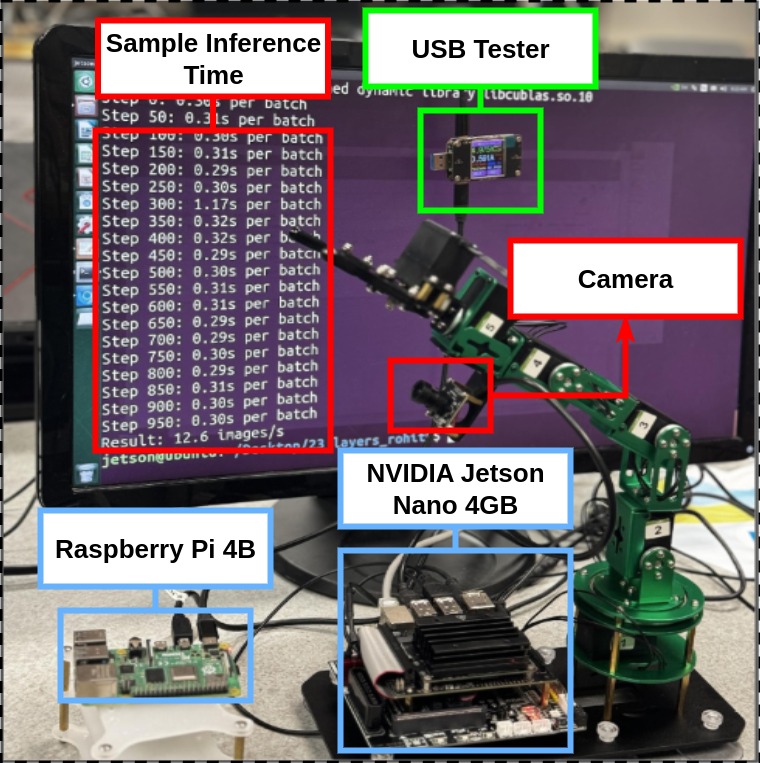}
\vspace{-0.5cm}
\caption[ ]
{Comprehensive edge-computing testbed integrating Raspberry Pi 4B and NVIDIA Jetson Nano boards with a Raspberry Pi camera and USB-based power analysis tools, enabling systematic evaluation of computational load, energy consumption, and real-time performance.}
\label{fig:hw_setup}
\vspace{-0.5cm}
\end{center}
\end{figure}


To evaluate the performance of the proposed event autoencoder-based adequately, we train a suite of SOTA neural network architectures using both SEFD~\cite{Islam_descriptor:2024} and EBCD~\cite{Joey_ebcd:2025} datasets, including the You Only Look Once (YOLO) family of models: YOLOv4~\cite{Bochkovskiy_yolov4:2020}, YOLOv7~\cite{Wang_yolov7:2022}, and YOLOv9~\cite{wang2024yolov9}, as well as EfficientDet-b0~\cite{Tan_efficientdet:2020}, MobileNet-v1~\cite{Howard_mobilenets:2017}, and YuNet~\cite{wu2023miryunet}. These models are characterized by their accurate detections, in-depth calculations, lightweight nature, and high-speed inference in visual detection applications, following the guidelines provided by their original articles. In this study, we evaluated five  empirical, event-generation thresholds ($T_h = 4, 8, 12, 16, 20$), where each threshold represents the minimum pixel-intensity variation required between consecutive frames or with respect to an initial frame. Each threshold is derived from the evaluated datasets.
The corresponding results of our experiments are quantitatively analyzed in Table~\ref{tab:benchmarks} and Table~\ref{tab:benchmark_ebcd}.

All computational experiments were conducted on a workstation equipped with an Intel Xeon(R) 20-core processor, 32 GB RAM, and an NVIDIA T1000 GPU (4 GB), operating on Ubuntu 20.04.6 LTS. 

To demonstrate the suitability of the proposed model for resource-constrained environments, the event autoencoder was deployed on both Raspberry Pi 4B and NVIDIA Jetson Nano devices. The embedded hardware configuration and setup are illustrated in Figure~\ref{fig:hw_setup}.




\begin{table*}[t]
\centering
\footnotesize
\setlength{\tabcolsep}{5pt}
\renewcommand{\arraystretch}{1.12}

\caption{
Performance comparison on the SEFD~\cite{Islam_descriptor:2024} test dataset under temporal thresholds $T_h \in {\{4, 8, 12, 16\}}$.
Our autoencoder classifier achieves competitive performance on frame-level data with orders of magnitude fewer parameters than heavyweight baselines.}

\label{tab:benchmarks}

\begin{tabular}{@{}l c c ccccc@{}}
\toprule
\hspace{1cm}\textbf{Model} &
\textbf{Params} &
\textbf{$T_h$} &
\textbf{Accuracy} &
\textbf{Precision} &
\textbf{Recall} &
\textbf{F1-Score} &
\textbf{FLOPs} \\
\midrule

\multirow{4}{*}{YOLOv4~\cite{Bochkovskiy_yolov4:2020}} &
\multirow{4}{*}{60.3M} & 
4 & 83.81 & 94.00 & 89.00 & 91.00 & \multirow{4}{*}{59.56B} \\
& & 8 & 80.33 & 89.00 & 89.00 & 89.00 \\
& & 12 & 79.98 & 91.00 & 86.00 & 89.00 \\
& & 16 & 74.49 & 93.00 & 79.00 & 85.00 \\
\midrule

\multirow{4}{*}{YOLOv7~\cite{Wang_yolov7:2022}} &
\multirow{4}{*}{25.2M} &
4 & 86.71 & 94.00 & 92.00 & 93.00 & \multirow{4}{*}{43.61B} \\
& & 8 & 82.95 & 92.00 & 90.00 & 91.00 \\
& & 12 & 83.44 & 95.00 & 87.00 & 91.00 \\
& & 16 & 81.88 & 95.00 & 85.00 & 90.00 \\
\midrule

\multirow{4}{*}{YOLOv9~\cite{wang2024yolov9}} &
\multirow{4}{*}{60.5M} &
4 & 97.69 & 90.20 & 93.90 & 92.01 & \multirow{4}{*}{263.9B} \\
& & 8  & 98.14 & 92.30 & 96.30 & 94.26 \\
& & 12 & 98.25 & 96.20 & 94.00 & 95.09 \\
& & 16 & 97.28 & 95.80 & 91.60 & 93.65 \\
\midrule

\multirow{4}{*}{EfficientDet-b0~\cite{Tan_efficientdet:2020}} &
\multirow{4}{*}{3.9M} &
4 & 94.55 & 97.68 & 96.72 & 97.21 & \multirow{4}{*}{4.87B} \\
& & 8  & 93.62 & 97.13 & 96.28 & 96.71 \\
& & 12 & 97.34 & 97.34 & 95.96 & 96.64 \\
& & 16 & 97.30 & 97.30 & 94.54 & 95.91 \\
\midrule

\multirow{4}{*}{MobileNet-v1~\cite{Howard_mobilenets:2017}} &
\multirow{4}{*}{6.05M} &
4 & 91.64 & 96.65 & 94.64 & 95.64 & \multirow{4}{*}{43.41B} \\
& & 8 & 89.44 & 92.64 & 96.28 & 94.43 \\
& & 12 & 92.11 & 96.36 & 95.41 & 95.88 \\
& & 16 & 89.11 & 96.01 & 98.14 & 97.06 \\
\midrule

\multirow{4}{*}{YuNet~\cite{wu2023miryunet}} &
\multirow{4}{*}{72.3K} &
4 & 85.46 & 93.00 & 85.00 & 89.00 & \multirow{4}{*}{486M} \\
& & 8 & 84.70 & 88.00 & 85.00 & 87.00 \\
& & 12 & 74.64 & 93.00 & 75.00 & 83.00 \\
& & 16 & 84.15 & 92.00 & 84.00 & 88.00 \\
\midrule

\multirow{4}{*}{\textbf{Autoencoder Classifier 100\% (Ours)}} &
\multirow{4}{*}{\textbf{1.7M}} &
\textbf{4} & \textbf{91.82} & \textbf{89.07} & \textbf{89.07} & \textbf{89.07} & \multirow{4}{*}{\textbf{6.50B}} \\
& & \textbf{8} & \textbf{93.01} & \textbf{88.27} & \textbf{93.77} & \textbf{90.94} \\
& & \textbf{12} & \textbf{92.15} & \textbf{86.70} & \textbf{93.33} & \textbf{89.89} \\
& & \textbf{16} & \textbf{90.02} & \textbf{82.80} & \textbf{92.57} & \textbf{87.41} \\
\midrule

\multirow{4}{*}{\textbf{Autoencoder Classifier 50\% (Ours)}} &
\multirow{4}{*}{\textbf{458K}} &
\textbf{4}  & \textbf{87.19} & \textbf{89.19} & \textbf{84.63} & \textbf{86.85} & \multirow{4}{*}{\textbf{1.66B}} \\
& & \textbf{8}  & \textbf{88.96} & \textbf{89.29} & \textbf{88.54} & \textbf{88.91} \\
& & \textbf{12} & \textbf{89.79} & \textbf{89.11} & \textbf{90.68} & \textbf{89.88} \\
& & \textbf{16} & \textbf{87.99} & \textbf{86.38} & \textbf{90.21} & \textbf{88.25} \\

\bottomrule
\end{tabular}
\end{table*}


\begin{table*}[t]
\centering
\footnotesize
\setlength{\tabcolsep}{5pt}
\renewcommand{\arraystretch}{1.12}

\caption{
Performance comparison on the EBCD~\cite{Joey_ebcd:2025} test dataset} across temporal thresholds $T_h \in \{12, 16, 20\}$.  
The proposed autoencoder classifier consistently outperforms prior models on frame-level data while using significantly fewer parameters.

\label{tab:benchmark_ebcd}

\begin{tabular}{@{}l c c ccccc@{}}
\toprule
\hspace{1cm}\textbf{Model} &
\textbf{Parameters} &
\textbf{$T_h$} &
\textbf{Accuracy} &
\textbf{Precision} &
\textbf{Recall} &
\textbf{F1-Score} &
\textbf{FLOPs} \\
\midrule

\multirow{3}{*}{YOLOv4~\cite{Bochkovskiy_yolov4:2020}} &
\multirow{3}{*}{60.3M} &
12 & 88.00 & 96.00 & 80.00 & 87.00 & \multirow{3}{*}{59.56B} \\
& & 16 & 87.50 & 94.00 & 81.00 & 87.00 \\
& & 20 & 89.00 & 94.00 & 84.00 & 87.00 \\
\midrule

\multirow{3}{*}{YOLOv7~\cite{Wang_yolov7:2022}} &
\multirow{3}{*}{25.2M} &
12 & 77.99 & 97.00 & 78.00 & 87.00 & \multirow{3}{*}{43.61B} \\
& & 16 & 77.99 & 97.00 & 78.00 & 86.00 \\
& & 20 & 79.35 & 97.00 & 79.00 & 87.00 \\
\midrule

\multirow{3}{*}{YOLOv9~\cite{wang2024yolov9}} &
\multirow{3}{*}{60.5M} &
12 & 98.75 & 97.80 & 93.50 & 95.60 & \multirow{3}{*}{263.9B} \\
& & 16 & 98.22 & 96.30 & 95.20 & 95.75 \\
& & 20 & 98.75 & 96.50 & 92.60 & 94.51 \\
\midrule

\multirow{3}{*}{EfficientDet-b0~\cite{Tan_efficientdet:2020}} &
\multirow{3}{*}{3.9M} &
12 & 58.22 & 98.44 & 58.22 & 73.17 &  \multirow{3}{*}{43.41B} \\
& & 16 & 57.03 & 99.21 & 57.03 & 72.43 \\
& & 20 & 46.40 & 100.00 & 46.40 & 63.90 \\
\midrule

\multirow{3}{*}{MobileNet-v1~\cite{Howard_mobilenets:2017}} &
\multirow{3}{*}{6.05M} &
12 & 26.12 & 98.30 & 26.20 & 41.40 & \multirow{3}{*}{43.41B} \\
& & 16 & 74.96 & 95.60 & 77.60 & 85.50 & \\
& & 20 & 35.96 & 94.10 & 36.80 & 52.90 \\
\midrule

\multirow{3}{*}{YuNet~\cite{wu2023miryunet}} &
\multirow{3}{*}{72.3K} &
12 & 84.16 & 93.00 & 84.00 & 88.00 & \multirow{3}{*}{486M} \\
& & 16 & 82.91 & 89.00 & 83.00 & 86.00 \\
& & 20 & 82.45 & 92.00 & 82.00 & 87.00 \\
\midrule

\multirow{3}{*}{\textbf{Autoencoder Classifier 100\% (Ours)}} &
\multirow{3}{*}{\textbf{1.7M}} &
\textbf{12} & \textbf{92.16} & \textbf{87.32} & \textbf{93.21} & \textbf{89.56} & \multirow{3}{*}{\textbf{13.87B}}\\
& & \textbf{16} & \textbf{93.47} & \textbf{88.04} & \textbf{92.77} & \textbf{90.09} \\
& & \textbf{20} & \textbf{94.38} & \textbf{88.69} & \textbf{94.45} & \textbf{91.13} \\
\midrule

\multirow{3}{*}{\textbf{Autoencoder Classifier 50\% (Ours)}} &
\multirow{3}{*}{\textbf{458K}} &
\textbf{12} & \textbf{88.79} & \textbf{82.47} & \textbf{87.36} & \textbf{84.84} & \multirow{3}{*}{\textbf{1.66B}} \\
& & \textbf{16} & \textbf{90.11} & \textbf{83.24} & \textbf{88.02} & \textbf{85.56} \\
& & \textbf{20} & \textbf{91.04} & \textbf{84.19} & \textbf{89.24} & \textbf{86.64} \\
\bottomrule

\end{tabular}
\end{table*}


\paragraph{Results and Discussion}
Compared to the existing SOTA models trained on the SEFD dataset, the proposed autoencoder classifier achieves higher accuracy than YOLOv4, YOLOv7, MobileNets-v, and YuNet. The YOLOv9 architecture achieves the best accuracy across both the SEFD and EBCD datasets compared to the competing models. The average accuracies of YOLOv9 across different thresholds are 97.84\% and 98.57\% for the SEFD and EBCD datasets, respectively. The proposed event autoencoder-based classifier has 2.3$\times$, 3.6$\times$, and 35.6$\times$ fewer parameters than EfficientDet-b0, MobileNet-v1, and YOLOv9 models, respectively.
Likewise, the proposed autoencoder classifier also expresses very high accuracy on the EBCD dataset. In Table~\ref{tab:benchmark_ebcd}, the model achieves an average accuracy of $93.33\%$ across the three thresholds: ($T_h = 12, 16, 20$), of the crossing dataset. The proposed model outshines YOLOv4, YOLOv7, EfficientDet-b0, MobileNet-v1, and YuNet, only being behind YOLOv9 in accuracy by $\approx 4 \text{-}6\%$.

For assessment, the event autoencoder was trained and tested on two categories of data: SEFD event samples representing positive instances and non-face samples serving as negative instances. 
The training dynamics of the proposed event autoencoder reveal strong convergence behavior, yielding an average training loss of 0.1593, as depicted in Figure~\ref{fig:loss_autoencoder}(a). After reducing the network capacity to 50\% of its original number of filters, the model attains an even lower average training loss of 0.0884, presented in Figure~\ref{fig:loss_autoencoder}(b). This outcome highlights the model’s scalability, demonstrating that a substantial reduction in filter count not only preserves but can enhance learning efficiency. Notably, compressing the architecture from full capacity to 50\% filters results in a fourfold reduction in overall model size, underscoring the resilience of the autoencoder design in achieving significant computational savings while sustaining high-quality feature learning and reconstruction performance.

\begin{figure}[t]
\begin{center}
\includegraphics[width = 0.5\textwidth]{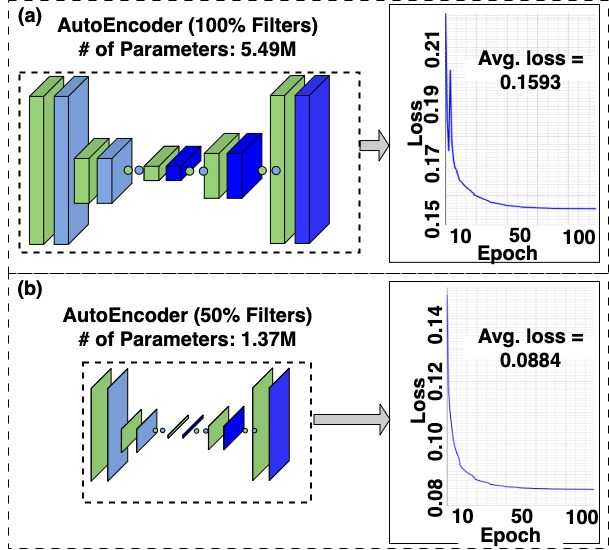}
\vspace{-0.7cm}
\caption[ ]
{(a) Training curve of the full-capacity event autoencoder, yielding an average loss close to fifteen hundredths. (b) Performance of the reduced 50\%–filter configuration, which achieves a lower average training loss close to one-tenth, illustrating the model’s resilience under substantial architectural compression.}
\label{fig:loss_autoencoder}
\vspace{-0.7cm}
\end{center}
\end{figure}



Accurate reconstruction is typically measured using a pixel-wise comparison, BCE analysis, or Structural Similarity Index Measure (SSIM)~\cite{WangSSIM2004} calculation, or a combination of all. Strong reconstruction fidelity indicates that the model successfully retains the essential structural and statistical properties of the initial data, thereby enhancing the quality of extracted features for subsequent tasks such as classification, segmentation, and denoising. In the context of outlier identification, improved reconstruction performance enables the system to more effectively differentiate between normal data patterns and aberrant inputs, thereby reducing error rates and increasing reliability.

\begin{figure*}[t]
		\begin{center}
			\vspace{-0.250cm}
			\includegraphics[width = \textwidth]{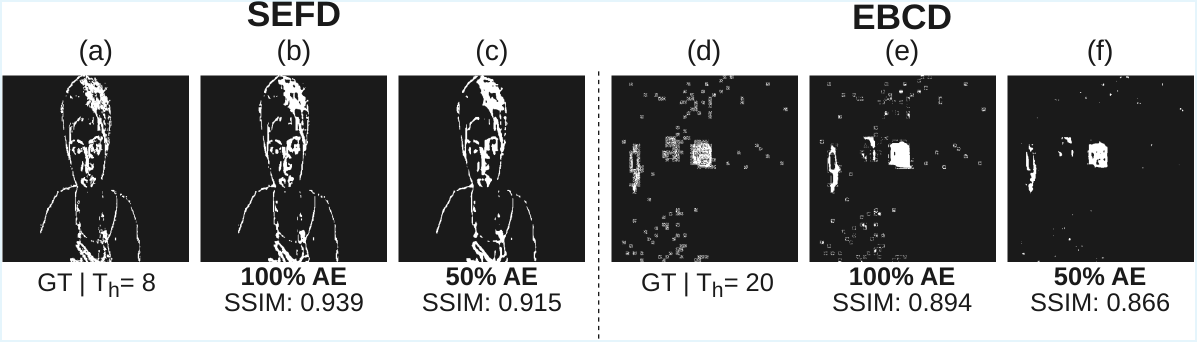}
			\vspace{-0.75cm}
			\caption{
            The reconstruction outputs of the proposed models exhibit excellent accuracy, comparable to their model sizes.} (a \& d) show the ground-truth (GT), pre-processed image, at $T_h=8$ and $T_h=20$ of the SEFD and EBCD datasets, respectively. (b \& e) represent the reconstructed outputs from the 100\% autoencoder with $\mathrm{SSIM} = 0.939$ and $\mathrm{SSIM} = 0.894$. (c \& f) are the reconstructed outputs from the 50\% autoencoder with $\mathrm{SSIM} = 0.915$ and $\mathrm{SSIM} = 0.866$.
				\label{fig:autoencoder_test}
			\vspace{-0.35cm}
		\end{center}
\end{figure*}

\begin{figure}[t!]
		\begin{center}
			\includegraphics[width = 0.489\textwidth]{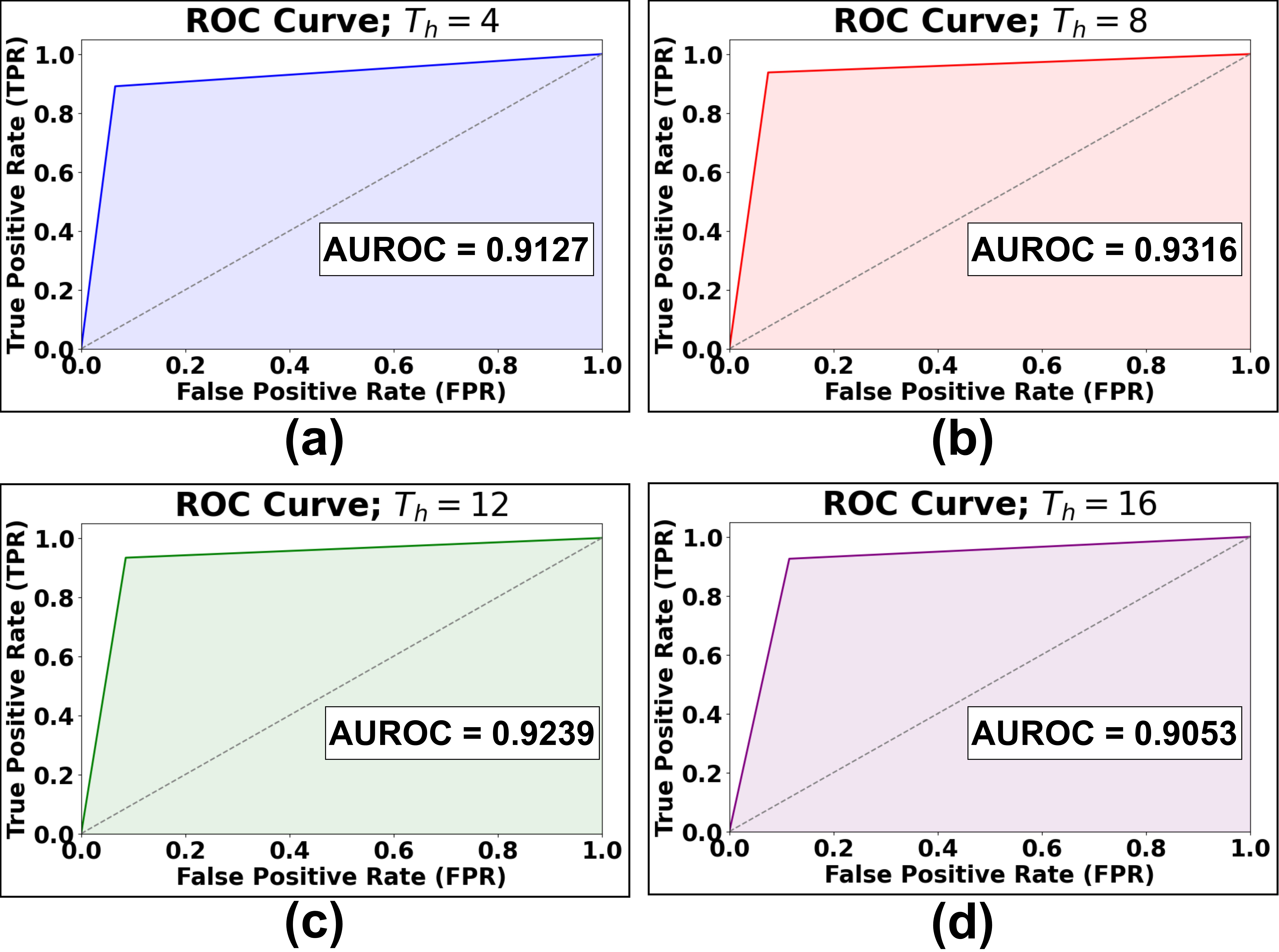}
			\vspace{-0.4cm}
			\caption {AUROC performance of the proposed autoencoder-based classifier across SEFD threshold settings. All configurations achieve AUC values above 90\%, with peak performance at $T_h = 8$}
				\label{fig:auroc}
			\vspace{-0.7cm}
		\end{center}
\end{figure}

Figure~\ref{fig:autoencoder_test} illustrates the reconstruction capability of the proposed event-driven autoencoder. The input event frame provided to the network is depicted in Figure~\ref{fig:autoencoder_test}(a). The full-capacity configuration (100\% filter model) achieves an SSIM of 0.939, as shown in Figure~\ref{fig:autoencoder_test}(b). By contrast, the reduced configuration utilizing only 50\% of the filters attains a slightly lower SSIM of 0.915, while achieving a 4$\times$ reduction in model size, as presented in Figure~\ref{fig:autoencoder_test}(c). These results highlight the effectiveness of the proposed design in maintaining high-quality reconstruction even under significant architectural compression. Likewise, with the EBCD dataset, the reconstructed SSIM of the 50\% autoencoder decreases by only 0.028 from that of the full-sized model in Figure~\ref{fig:autoencoder_test}(f).

\begin{figure}[b]
		\begin{center}
			\includegraphics[width = 0.485\textwidth]{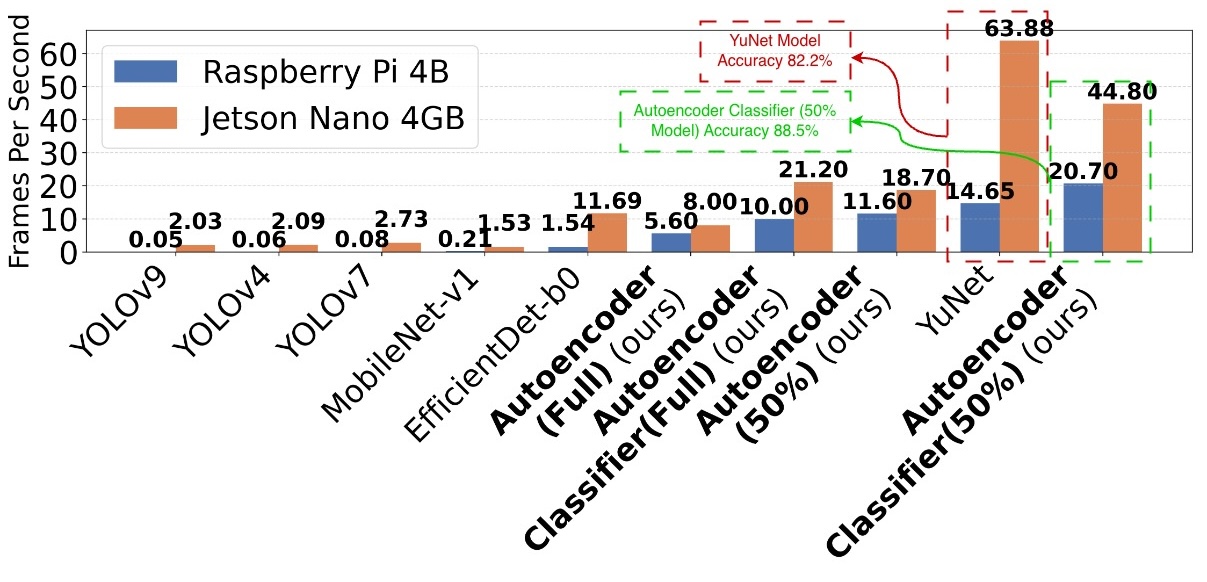}
			\vspace{-0.4cm}
			\caption {Runtime evaluation of the proposed architectures compared with contemporary work on Raspberry Pi 4B and NVIDIA Jetson Nano (4GB). The results show reduced execution time and higher FPS, with the autoencoder classifier achieving 22.1$\times$ the FPS of YOLOv9. Although YuNet is fastest, the proposed 50\% filter classifier is 6.3\% more accurate.}
				\label{fig:hw_results}
		\end{center}
\end{figure}

\begin{figure*}[t]
\begin{center}
\vspace{-0cm}
\includegraphics[width=0.75\textwidth]{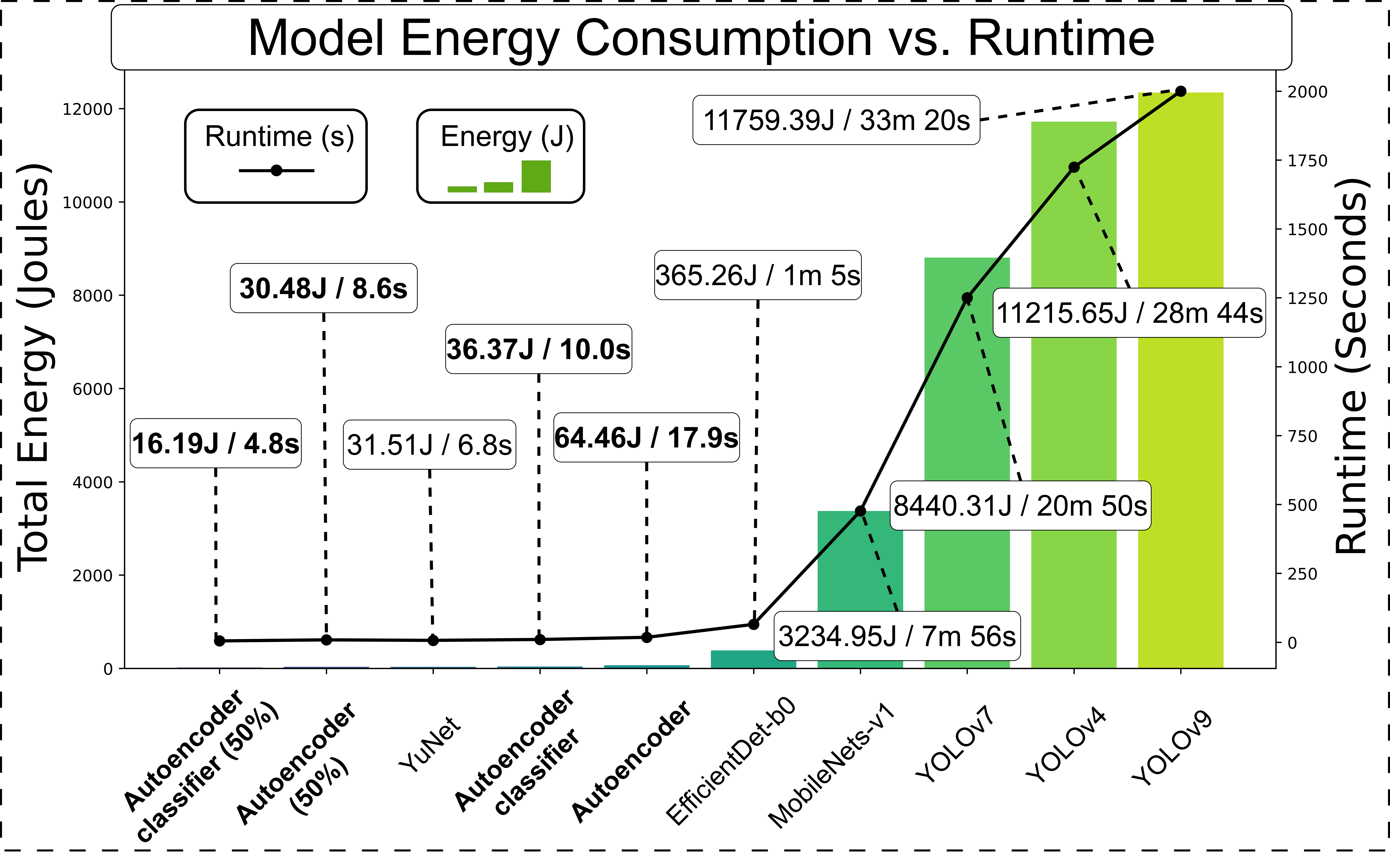}
\vspace{-0.3cm}
\caption[ ]
{
Energy consumption and runtime during inference on a Raspberry Pi 4B.
The proposed 50\% Autoencoder Classifier and 50\% Autoencoder require only 16.19~J and 30.48~J to process 100 images, achieving more than a $2\times$ energy reduction over their full-size counterparts (36.37~J and 64.46~J). Their runtimes are similarly low at 4.8~s and 8.6~s. In contrast, lightweight baselines such as MobileNet-v1 and YOLOv7 consume 3234.95~J and 8440.31~J (476.2~s and 1250.0~s), while the most demanding model, YOLOv9, reaches 11,759.39~J and 2000.0~s—over $700\times$ the energy usage of the 50\% Autoencoder Classifier.}
\label{fig:energy-runtime}
\vspace{-0.5cm}
\end{center}
\end{figure*}

Performance assessment of binary classification systems frequently incorporates the Receiver Operating Characteristic (ROC) curve, a foundational analytical tool in machine learning and anomaly detection~\cite{Islam_ROC:2022}. The ROC curve depicts the relationship between the sensitivity (true positive proportion) and the false positive proportion, thereby illustrating how effectively a model separates positive samples from negative ones across varying decision thresholds~\cite{Fawcett:2006}. The Area Under the ROC Curve (AUROC) provides a scalar summary of this relationship; an AUROC value of 1.0 indicates ideal discrimination capability, whereas a value of 0.5 signifies performance no better than random guessing~\cite{Hanley:1982}. Owing to its threshold-invariant nature, AUROC is routinely employed in applications such as outlier identification 
to benchmark and refine classifier performance.

Figure~\ref{fig:auroc} summarizes the AUROC performance of the proposed autoencoder-based classifier across all SEFD event-generation thresholds, $T_h$. At the lowest threshold ($T_h = 4$), where event activity is highest, the classifier maintains strong separation between event and non-event samples despite the increased pixel-level density. Performance peaks at $T_h = 8$, yielding the highest AUROC and indicating optimal discriminative capability. Although AUROC values decrease slightly at higher thresholds ($T_h = 12$ and $T_h = 16$) due to sparser event representations, the classifier retains consistently high performance across all settings, demonstrating robust detection capability under varying activity levels.

\paragraph{Performance on Embedded Platforms}
When transferring models to the embedded devices, all were evaluated using 32FP (32-bit floating point). The results of the embedded platform analysis can be depicted in  Figure~\ref{fig:hw_results}. On the Raspberry Pi 4B platform, the 50\%–filter configuration of the proposed event autoencoder delivers substantial speed gains, achieving a $2.71\times$ increase in FPS over the full model and a $55.24\times$ improvement relative to MobileNet-v1. Using the same hardware platform, the 50\% event-autoencoder classifier demonstrates similarly strong performance, operating more than double the FPS of its 100\% counterpart, and nearly $100\times$ the throughput of MobileNet-v1. Furthermore, the classifier equipped with the 50\% filter reduction outperforms EfficientDet-b0 by a factor of $13.44\times$ in terms of inference speed. The 50\% autoencoder classifier also achieves $1.41\times$ the FPS of the tiny YuNet model. Considering the large size of the YOLO-family models, the proposed model exhibits up to $414\times$ more FPS. 

As mentioned, the evaluation was extended to the NVIDIA Jetson Nano platform to further assess the efficiency of the proposed architecture on an embedded GPU. The event autoencoder configured with a 50\% filter reduction achieves significant throughput advantages, operating at nearly 9$\times$ the FPS of YOLOv4~\cite{Bochkovskiy_yolov4:2020} and about 37$\times$ the FPS of MobileNet-v1~\cite{Howard_mobilenets:2017}. The corresponding event-based classifier demonstrates even greater performance gains, delivering 21.4 times higher FPS than YOLOv4 and nearly 88$\times$ higher FPS than MobileNet-v1, confirming the model’s suitability for real-time processing on low-power embedded hardware. However, it can be noted that with the increase in processing power, YuNet~\cite{wu2023miryunet} exhibits the fastest runtimes.


\paragraph{Energy Consumption}
The importance of energy efficient computations are well studied in the literature~\cite{khwa2025mixed, Islam_cmcs:2017, Tossoun:2025, Islam_HCDN:2019, ambrogio2023analog, Islam_iscas:2014, xi2026cnn, Islam_cjece:2019, vimala2025study, Islam_asicon:2011, rajendra2025ultra, Islam_neg:2021}.
Hence, we evaluated the energy consumption of all tested models by measuring their power usage during sequential inference on 100 images. To ensure fair energy measurement, the proposed model and all SOTA models use the same 100 test images. We ran the same experiments 10 times and averaged the results for each data point. Figure~\ref{fig:energy-runtime} reports the total energy usage (in joules) and runtime (in seconds) for each model when deployed on a Raspberry Pi~4B. Measurements were collected using an inline USB power meter that sampled real-time current (\(I\)) and voltage (\(V\)) throughout execution. The resultant energy values are subtracted from the idle power consumption, recorded at 0.2929525 watts (joules per second).

To reduce noise and isolate stable behavior, the recorded power trace was partitioned into three operational phases: warmup, steady-state inference, and cooldown, with respective durations \(w_1, w_2, w_3\). For each phase, average current and voltage were computed independently as \(\overline{I}_k\) and \(\overline{V}_k\) for \(k \in \{1,2,3\}\). Phase-wise average power was then obtained as
\begin{equation}
\overline{P}_k = \overline{V}_k \cdot \overline{I}_k.
\end{equation}
The overall average power was estimated through a weighted temporal mean,
\begin{equation}
\text{Avg}(P) =
\frac{
w_1\,\overline{P}_1 +
w_2\,\overline{P}_2 +
w_3\,\overline{P}_3
}{
w_1 + w_2 + w_3
}.
\label{eq:avg_power_corrected}
\end{equation}
Here we approximate the time integral of instantaneous power over the entire execution window. Total energy consumption was then computed as
\begin{equation}
E = \text{Avg}(P)\times t.
\label{eq:energy_final}
\end{equation}
The total inference time across all 100 processed images is represented as $t$. This weighted approach suppresses transient fluctuations, and emphasizes the dominant steady-state inference activity.

Across all evaluated models, our proposed architectures (bolded in Figure~\ref{fig:energy-runtime}) exhibit substantial gains in energy efficiency. The 50\% autoencoder classifier and 50\% autoencoder consumed only 16.19~J and 30.48~J, respectively—representing dramatic reductions relative to commonly deployed SOTA detectors. Specifically, the 50\% autoencoder classifier is approximately \(726.3\times\), \(199.8\times\), \(22.6\times\), and \(1.95\times\) more energy efficient than YOLOv9, MobileNet-v1, EfficientDet-b0, and YuNet, respectively. Although highly accurate in object classification, the YOLO-family models incur significantly larger energy costs due to their deep and computationally intensive architectures.
\paragraph{Encoder size ablation analysis}
To further evaluate the scalability of the proposed architecture, we examine the effect of further reducing the encoder capacity. In addition to the 50\% model, we evaluate another reduced variant obtained by progressively removing convolutional filters, namely, a 25\% configuration. As the encoder size decreases, the number of parameters and FLOPs are substantially reduced while maintaining competitive recognition performance. In Table~\ref{tab:ablation_sefd}, the 25\% model reduces the architecture to only 132K parameters and 429M FLOPs, while still achieving strong classification performance across the evaluated threshold settings. This behavior further highlights the robustness of the learned representation and suggests that the computational savings primarily stem from the lightweight encoder design.

\begin{table}[h]
\centering
\footnotesize
\setlength{\tabcolsep}{5pt}
\renewcommand{\arraystretch}{1.12}

\caption{Ablation on the analysis of a further 
reduced autoencoder classifier, evaluated on the SEFD~\cite{Islam_descriptor:2024} dataset across temporal thresholds $T_h \in \{4, 8, 12, 16\}$.
}
\label{tab:ablation_sefd}
\resizebox{\columnwidth}{!}{%
\begin{tabular}{@{}l c c ccccc@{}}
\toprule
\hspace{1cm}\textbf{Model} &
{Parameters} &
{$T_h$} &
{Accuracy} &
{Precision} &
{Recall} &
{F1-Score} &
{FLOPs} \\
\midrule
\multirow{4}{*}{{Autoencoder Classifier 25\% (Ours)}} &
\multirow{4}{*}{{132K}} &
{4} & {88.79} & {86.48} & {82.97} & {84.69} & \multirow{4}{*}{{429M}} \\
& & {8} & {86.30} & {86.33} & {86.25} & {86.29} \\
& & {12} & {86.46} & {85.11} & {88.39} & {86.71} \\
& & {16} & {84.56} & {82.32} & {88.02} & {85.07} \\
\bottomrule

\end{tabular}%
}
\end{table}

For additional architectural context, reduced detector variants remain comparatively heavier. For example, when exploring the yolo-family models: YOLOv4-tiny~\cite{Bochkovskiy_yolov4:2020} uses 5.6M parameters and 5.80B FLOPs, YOLOv7-tiny~\cite{Wang_yolov7:2022} uses 3.9M parameters and 6.79B FLOPs, and YOLOv9-tiny~\cite{wang2024yolov9} uses 2.6M parameters and 10.70B FLOPs. By comparison, even the full proposed model remains compact, while the reduced 25\% configuration further lowers complexity drastically.


\section{Conclusion}





The architecture presented herein delivers a deployable event-driven autoencoder with an integrated classifier optimized for real-time execution on power-constrained vision hardware. The proposed system demonstrates robust compression and reconstruction of event streams, preserving critical spatiotemporal features while minimizing computational and energy overhead. Evaluations on the SEFD~\cite{Islam_descriptor:2024} and EBCD~\cite{Joey_ebcd:2025} datasets demonstrate classification accuracies of 93\% across multiple thresholds, outperforming YOLOv4~\cite{Bochkovskiy_yolov4:2020}, YOLOv7~\cite{Wang_yolov7:2022}, MobileNet-v1~\cite{Howard_mobilenets:2017}, and YuNet~\cite{wu2023miryunet}, and maintaining competitive accuracy within 4--6\% of YOLOv9~\cite{wang2024yolov9}, despite using 35.6$\times$ fewer parameters. The reconstruction pipeline achieves 99.97\% accuracy with the full model and retains 93.54\% accuracy even when scaled to 50\% of the filters, confirming the model’s resilience under aggressive parameter reduction. Implementation on embedded CPU (i.e., Raspberry Pi 4B) and GPU (i.e., NVIDIA Jetson Nano) platforms further validates the effectiveness of the proposed solution, achieving up to 87.84$\times$ higher FPS than MobileNet-v1~\cite{Howard_mobilenets:2017} and over 700$\times$ lower energy consumption than YOLOv9~\cite{wang2024yolov9}. The system delivers 20.7 FPS on Raspberry Pi and 44.8 FPS on Jetson Nano with the 50\% classifier model, while consuming only 16.19~J per 100 inferences. These results highlight the suitability of the proposed architecture for high-speed, low-power edge intelligence in consumer electronics. 
Moreover, this work provides a strong foundation for extending the proposed architecture to more complex multi-object and multi-class event datasets. Building on this basis, future work will explore richer event-based perception scenarios for real-time detection.



\section*{Acknowledgment}

We thank S.R.S.K. Tummala of UMBC for his early work on the early architecture and results. We also extend our appreciation to R. Kankipati, R. Robucci (UMBC), and C. Howard (Oculi.ai) for their insightful assistance with analysis and discussion.


\bibliographystyle{IEEEtran}
\bibliography{main}

@inproceedings{wang2022ev,
  title={{Ev-catcher: High-speed object catching using low-latency event-based neural networks}},
  author={Wang, Zhen and Cladera, Francisco and Bisulco, Adam and Lee, Dongheui},
  booktitle={IEEE International Conference on Robotics and Automation (ICRA)},
  pages={1045--1051},
  year={2022},
  organization={IEEE}
}

@ARTICLE{Islam_tcasii:2021,
  author={Islam, Riadul and Saha, Biprangshu and Bezzam, Ignatius},
  journal={IEEE Transactions on Circuits and Systems II: Express Briefs}, 
  title={{Resonant Energy Recycling SRAM Architecture}}, 
  year={2021},
  volume={68},
  number={4},
  pages={1383-1387},
  doi={10.1109/TCSII.2020.3029203}
  }

@article{islam2024benchmarking,
  title={{Benchmarking Artificial Neural Network Architectures for High-Performance Spiking Neural Networks}},
  author={Islam, Riadul and Majurski, Patrick and Kwon, Jun and Sharma, Anurag and Tummala, Sri Ranga Sai Krishna},
  journal={Sensors},
  volume={24},
  number={4},
  pages={1329},
  year={2024},
  publisher={MDPI}
}

@inproceedings{kamata2022fully,
  title={{Fully spiking variational autoencoder}},
  author={Kamata, Hiromichi and Mukuta, Yusuke and Harada, Tatsuya},
  booktitle={Proceedings of the AAAI conference on artificial intelligence},
  volume={36},
  number={6},
  pages={7059--7067},
  year={2022}
}

@inproceedings{gruel2022event,
  title={{Event data downscaling for embedded computer vision}},
  author={Gruel, Adrien and Martinet, Johann and Serrano-Gotarredona, Teresa and others},
  booktitle={International Conference on Computer Vision Systems},
  pages={1--6},
  year={2022}
}

@article{Tai_pottsmgnet:2024,
  title={{PottsMGNet: A mathematical explanation of encoder-decoder based neural networks}},
  author={Tai, Xue-Cheng and Liu, Hao and Chan, Raymond},
  journal={SIAM Journal on Imaging Sciences},
  volume={17},
  number={1},
  pages={540--594},
  year={2024},
  publisher={SIAM}
}

@INPROCEEDINGS{Chen_auto_cvpr:2023,
  author={Chen, Anthony and Zhang, Kevin and Zhang, Renrui and Wang, Zihan and Lu, Yuheng and Guo, Yandong and Zhang, Shanghang},
  booktitle={2023 IEEE/CVF Conference on Computer Vision and Pattern Recognition (CVPR)}, 
  title={Pi{MAE}: Point Cloud and Image Interactive Masked Autoencoders for 3{D} Object Detection}, 
  year={2023},
  volume={},
  number={},
  pages={5291-5301},
  doi={10.1109/CVPR52729.2023.00512}
}

@article{Wang_auto:2016,
title = {Auto-encoder based dimensionality reduction},
journal = {Neurocomputing},
volume = {184},
pages = {232-242},
year = {2016},
issn = {0925-2312},
doi = {https://doi.org/10.1016/j.neucom.2015.08.104},
url = {https://www.sciencedirect.com/science/article/pii/S0925231215017671},
author = {Yasi Wang and Hongxun Yao and Sicheng Zhao},
}

@article{Wan_event_auto:2022,
   title={Learning Dense and Continuous Optical Flow From an Event Camera},
   volume={31},
   ISSN={1941-0042},
   url={http://dx.doi.org/10.1109/TIP.2022.3220938},
   DOI={10.1109/tip.2022.3220938},
   journal={IEEE Transactions on Image Processing},
   publisher={Institute of Electrical and Electronics Engineers (IEEE)},
   author={Wan, Zhexiong and Dai, Yuchao and Mao, Yuxin},
   year={2022},
   pages={7237--7251} 
}

@inproceedings{Hidalgo_rnn_auto:2020,
  title={Learning monocular dense depth from events},
  author={Hidalgo-Carrio, Javier and Gehrig, Daniel and Scaramuzza, Davide},
  booktitle={International Conference on 3D Vision (3DV)},
  pages={534--542},
  year={2020},
  organization={IEEE}
}

@article{Gehrig_rnn_auto:2021,
  title={Combining events and frames using recurrent asynchronous multimodal networks for monocular depth prediction},
  author={Gehrig, Daniel and Ruegg, Michelle and Gehrig, Mathias and Hidalgo-Carrio, Javier and Scaramuzza, Davide},
  journal={IEEE Robotics and Automation Letters},
  volume={6},
  number={2},
  pages={2822--2829},
  year={2021},
  publisher={IEEE}
}

@misc{Zhang_auto_fusion:2021,
      title={ISSAFE: Improving Semantic Segmentation in Accidents by Fusing Event-based Data}, 
      author={Jiaming Zhang and Kailun Yang and Rainer Stiefelhagen},
      year={2021},
      eprint={2008.08974},
      archivePrefix={arXiv},
      primaryClass={cs.CV}
}

@INPROCEEDINGS{Han_auto_fusion:2020,
  author={Han, Jin and Zhou, Chu and Duan, Peiqi and Tang, Yehui and Xu, Chang and Xu, Chao and Huang, Tiejun and Shi, Boxin},
  booktitle={2020 IEEE/CVF Conference on Computer Vision and Pattern Recognition (CVPR)}, 
  title={Neuromorphic Camera Guided High Dynamic Range Imaging}, 
  year={2020},
  volume={},
  number={},
  pages={1727-1736},
  doi={10.1109/CVPR42600.2020.00180}
}

@ARTICLE{Li_rnn_auto:2020,
  author={Li, Hui and Ma, Kede and Yong, Hongwei and Zhang, Lei},
  journal={IEEE Transactions on Image Processing}, 
  title={Fast Multi-Scale Structural Patch Decomposition for Multi-Exposure Image Fusion}, 
  year={2020},
  volume={29},
  number={},
  pages={5805-5816},
  doi={10.1109/TIP.2020.2987133}
}

@article{li2023sodformer,
  title={{Sodformer: Streaming object detection with transformer using events and frames}},
  author={Li, Daming and Tian, Yuchao and Li, Jianhua},
  journal={IEEE Transactions on Pattern Analysis and Machine Intelligence},
  year={2023}
}

@article{li2022asynchronous,
  title={{Asynchronous spatio-temporal memory network for continuous event-based object detection}},
  author={Li, Jianhua and Zhu, Ling and Xiang, Xiaobo and Huang, Tao},
  journal={IEEE Transactions on Neural Networks and Learning Systems},
  year={2022}
}

@ARTICLE{Yu_spikingvit:2025,
  author={Yu, Lixing and Chen, Hanqi and Wang, Ziming and Zhan, Shaojie and Shao, Jiankun and Liu, Qingjie and Xu, Shu},
  journal={{IEEE Transactions on Cognitive and Developmental Systems}}, 
  title={SpikingViT: A Multiscale Spiking Vision Transformer Model for Event-Based Object Detection}, 
  year={2025},
  volume={17},
  number={1},
  pages={130-146},
  doi={10.1109/TCDS.2024.3422873}}

@inproceedings{kim2022ev,
  title={{Ev-TTA: Test-time adaptation for event-based object recognition}},
  author={Kim, Jaeyoung and Hwang, Inho and Kim, Youngmin},
  booktitle={Proceedings of the IEEE/CVF Conference on Computer Vision and Pattern Recognition},
  pages={8769--8778},
  year={2022}
}

@inproceedings{zhang2022spiking,
  title={{Spiking transformers for event-based single object tracking}},
  author={Zhang, Jie and Dong, Bo and Zhang, Hanyu and Ding, Jiayuan and others},
  booktitle={Proceedings of the IEEE/CVF Conference on Computer Vision and Pattern Recognition},
  pages={11240--11250},
  year={2022}
}

@ARTICLE{OrchardNMNSIT2015,  
AUTHOR={Orchard, Garrick  and Jayawant, Ajinkya  and Cohen, Gregory K.  and Thakor, Nitish },       
TITLE={{Converting Static Image Datasets to Spiking Neuromorphic Datasets Using Saccades}},       
JOURNAL={Frontiers in Neuroscience},        
VOLUME={Volume 9 - 2015},
YEAR={2015},
URL={https://www.frontiersin.org/journals/neuroscience/articles/10.3389/fnins.2015.00437},
DOI={10.3389/fnins.2015.00437},
ISSN={1662-453X}
}

@INPROCEEDINGS{AmirDVSGesture2017,
  author={Amir, Arnon and Taba, Brian and Berg, David and Melano, Timothy and McKinstry, Jeffrey and Di Nolfo, Carmelo and Nayak, Tapan and Andreopoulos, Alexander and Garreau, Guillaume and Mendoza, Marcela and Kusnitz, Jeff and Debole, Michael and Esser, Steve and Delbruck, Tobi and Flickner, Myron and Modha, Dharmendra},
  booktitle={IEEE Conference on Computer Vision and Pattern Recognition (CVPR)}, 
  title={{A Low Power, Fully Event-Based Gesture Recognition System}}, 
  year={2017},
  volume={},
  number={},
  pages={7388-7397},
  doi={10.1109/CVPR.2017.781}
}

@inproceedings{lin2014microsoft,
  title={Microsoft coco: Common objects in context},
  author={Lin, Tsung-Yi and Maire, Michael and Belongie, Serge and Hays, James and Perona, Pietro and Ramanan, Deva and Doll{\'a}r, Piotr and Zitnick, C Lawrence},
  booktitle={European conference on computer vision},
  pages={740--755},
  year={2014},
  organization={Springer}
}

@article{DBLP:journals/corr/OrchardJCT15,
  author       = {Garrick Orchard and
                  Ajinkya Jayawant and
                  Gregory Cohen and
                  Nitish V. Thakor},
  title        = {Converting Static Image Datasets to Spiking Neuromorphic Datasets
                  Using Saccades},
  journal      = {CoRR},
  volume       = {abs/1507.07629},
  year         = {2015},
  url          = {http://arxiv.org/abs/1507.07629},
  eprinttype    = {arXiv},
  eprint       = {1507.07629},
  bibsource    = {dblp computer science bibliography, https://dblp.org}
}

@article{Reddy_recognizing:2013,
  title={{Recognizing 50 human action categories of web videos}},
  author={Reddy, Kishore K and Shah, Mubarak},
  journal={Machine vision and applications},
  volume={24},
  number={5},
  pages={971--981},
  year={2013},
  publisher={Springer}
}

@article{Binas_ddd17:2017,
  title={{DDD17: End-to-end DAVIS driving dataset}},
  author={Binas, Jonathan and Neil, Daniel and Liu, Shih-Chii and Delbruck, Tobi},
  journal={arXiv preprint arXiv:1711.01458},
  year={2017}
}

@article{Zhu_flow:2018,
   title={{The Multivehicle Stereo Event Camera Dataset: An Event Camera Dataset for 3D Perception}},
   volume={3},
   ISSN={2377-3774},
   url={http://dx.doi.org/10.1109/LRA.2018.2800793},
   DOI={10.1109/lra.2018.2800793},
   number={3},
   journal={IEEE Robotics and Automation Letters},
   publisher={Institute of Electrical and Electronics Engineers (IEEE)},
   author={Zhu, Alex Zihao and Thakur, Dinesh and Ozaslan, Tolga and Pfrommer, Bernd and Kumar, Vijay and Daniilidis, Kostas},
   year={2018},
   month=jul, pages={2032–2039} 
}

@misc{litmaps,
  title = {Litmaps | Your Literature Review Assistant},
  url = {https://litmaps.com},
  urldate = {2025-07-31},
  year = {2025},
  organization = {Litmaps.com}
}

@inproceedings{Bi_graph:2019,
  title={{Graph-based object classification for neuromorphic vision sensing}},
  author={Bi, Yin and Chadha, Aaron and Abbas, Alhabib and Bourtsoulatze, Eirina and Andreopoulos, Yiannis},
  booktitle={Proceedings of the IEEE/CVF international conference on computer vision},
  pages={491--501},
  year={2019}
}

@INPROCEEDINGS{Amir_gesture:2017,
  author={Amir, Arnon and Taba, Brian and Berg, David and Melano, Timothy and McKinstry, Jeffrey and Di Nolfo, Carmelo and Nayak, Tapan and Andreopoulos, Alexander and Garreau, Guillaume and Mendoza, Marcela and Kusnitz, Jeff and Debole, Michael and Esser, Steve and Delbruck, Tobi and Flickner, Myron and Modha, Dharmendra},
  booktitle={IEEE Conference on Computer Vision and Pattern Recognition (CVPR)}, 
  title={{A Low Power, Fully Event-Based Gesture Recognition System}}, 
  year={2017},
  volume={},
  number={},
  pages={7388-7397},
  doi={10.1109/CVPR.2017.781}
}

@article{Orchard_NCaltec:2015,
  title={{Converting static image datasets to spiking neuromorphic datasets using saccades}},
  author={Orchard, Garrick and Jayawant, Ajinkya and Cohen, Gregory K and Thakor, Nitish},
  journal={Frontiers in neuroscience},
  volume={9},
  pages={437},
  year={2015},
  publisher={Frontiers Media SA}
}

@article{Miao_neuromorphic:2019,
  title={{Neuromorphic vision datasets for pedestrian detection, action recognition, and fall detection}},
  author={Miao, Shu and Chen, Guang and Ning, Xiangyu and Zi, Yang and Ren, Kejia and Bing, Zhenshan and Knoll, Alois},
  journal={Frontiers in neurorobotics},
  volume={13},
  pages={38},
  year={2019},
  publisher={Frontiers Media SA}
}

@INPROCEEDINGS{dvs_gesture,
  author={Amir, Arnon and Taba, Brian and Berg, David and Melano, Timothy and McKinstry, Jeffrey and Di Nolfo, Carmelo and Nayak, Tapan and Andreopoulos, Alexander and Garreau, Guillaume and Mendoza, Marcela and Kusnitz, Jeff and Debole, Michael and Esser, Steve and Delbruck, Tobi and Flickner, Myron and Modha, Dharmendra},
  booktitle={2017 IEEE Conference on Computer Vision and Pattern Recognition (CVPR)}, 
  title={{A Low Power, Fully Event-Based Gesture Recognition System}}, 
  year={2017},
  volume={},
  number={},
  pages={7388-7397},
  doi={10.1109/CVPR.2017.781}}

@misc{prophesee_1Mpx:2020,
      title={{Learning to Detect Objects with a 1 Megapixel Event Camera}}, 
      author={Etienne Perot and Pierre de Tournemire and Davide Nitti and Jonathan Masci and Amos Sironi},
      year={2020},
      eprint={2009.13436},
      archivePrefix={arXiv},
      primaryClass={cs.CV},
      url={https://arxiv.org/abs/2009.13436}, 
}

@misc{prophesee_gen1:2020,
      title={{A Large Scale Event-based Detection Dataset for Automotive}}, 
      author={Pierre de Tournemire and Davide Nitti and Etienne Perot and Davide Migliore and Amos Sironi},
      year={2020},
      eprint={2001.08499},
      archivePrefix={arXiv},
      primaryClass={cs.CV},
      url={https://arxiv.org/abs/2001.08499}, 
}

@ARTICLE{Joey_ebcd:2025,
  author={Mulé, Joey and Challagundla, Dhandeep and Saini, Rachit and Islam, Riadul},
  journal={IEEE Data Descriptions}, 
  title={{Descriptor: Event-Based Crossing Dataset (EBCD)}}, 
  year={2025},
  volume={2},
  number={},
  pages={71-81},
  doi={10.1109/IEEEDATA.2025.3561760}}

@article{Islam_descriptor:2024,
  title={{Descriptor: Smart Event Face Dataset (SEFD)}},
  author={Islam, Riadul and Tummala, Sri Ranga Sai Krishna and Mul{\'e}, Joey and Kankipati, Rohith and Jalapally, Suraj and Challagundla, Dhandeep and Howard, Chad and Robucci, Ryan},
  journal={IEEE Data Descriptions},
  year={2024},
  publisher={IEEE}
}

@article{Kollias:2019,
author = {Kollias, Dimitrios and Tzirakis, Panagiotis and Nicolaou, Mihalis and Papaioannou, Athanasios and Zhao, Guoying and Schuller, Björn and Kotsia, Irene and Zafeiriou, Stefanos},
year = {2019},
month = {06},
pages = {},
title = {{Deep Affect Prediction in-the-Wild: Aff-Wild Database and Challenge, Deep Architectures, and Beyond}},
volume = {127},
journal = {International Journal of Computer Vision},
doi = {10.1007/s11263-019-01158-4}
}

@article{wu2023miryunet,
	title     = {YuNet: A Tiny Millisecond-level Face Detector},
	author    = {Wu, Wei and Peng, Hanyang and Yu, Shiqi},
	journal   = {Machine Intelligence Research},
	pages     = {1--10},
	year      = {2023},
	doi       = {10.1007/s11633-023-1423-y},
	publisher = {Springer}
}

@article{wang2024yolov9,
  title={{YOLOv9}: Learning What You Want to Learn Using Programmable Gradient Information},
  author={Wang, Chien-Yao  and Liao, Hong-Yuan Mark},
  booktitle={arXiv preprint arXiv:2402.13616},
  year={2024}
}

@inproceedings{zubic2023chaos,
  title={From chaos comes order: Ordering event representations for object recognition and detection},
  author={Zubi{\'c}, Nikola and Gehrig, Daniel and Gehrig, Mathias and Scaramuzza, Davide},
  booktitle={Proceedings of the IEEE/CVF International Conference on Computer Vision},
  pages={12846--12856},
  year={2023}
}

@inproceedings{torbunov2025evrt,
  title={Evrt-detr: Latent space adaptation of image detectors for event-based vision},
  author={Torbunov, Dmitrii and Ren, Yihui and Ghose, Animesh and Dim, Odera and Cui, Yonggang},
  booktitle={Proceedings of the IEEE/CVF International Conference on Computer Vision},
  pages={9812--9821},
  year={2025}
}

@misc{Tan_efficientdet:2020,
      title={{Efficient{D}et: Scalable and Efficient Object Detection}}, 
      author={Mingxing Tan and Ruoming Pang and Quoc V. Le},
      year={2020},
      eprint={1911.09070},
      archivePrefix={arXiv},
      primaryClass={cs.CV}
}

@misc{Howard_mobilenets:2017,
      title={{MobileNets: Efficient Convolutional Neural Networks for Mobile Vision Applications}}, 
      author={Andrew G. Howard and Menglong Zhu and Bo Chen and Dmitry Kalenichenko and Weijun Wang and Tobias Weyand and Marco Andreetto and Hartwig Adam},
      year={2017},
      eprint={1704.04861},
      archivePrefix={arXiv},
      primaryClass={cs.CV}
}

@article{Bochkovskiy_yolov4:2020,
  title={{Yolov4: Optimal speed and accuracy of object detection}},
  author={Bochkovskiy, Alexey and Wang, Chien-Yao and Liao, Hong-Yuan Mark},
  journal={arXiv preprint arXiv:2004.10934},
  year={2020},
  pages={1--17},
  doi={10.48550/arXiv.2004.10934}
}

@inproceedings{Wang_yolov7:2022,
  title={{YOLOv7: Trainable bag-of-freebies sets new state-of-the-art for real-time object detectors}},
  author={Wang, Chien-Yao and Bochkovskiy, Alexey and Liao, Hong-Yuan Mark},
  booktitle={Proceedings of the IEEE/CVF conference on computer vision and pattern recognition},
  pages={7464--7475},
  year={2023},
  doi={10.48550/arXiv.2207.02696}
}

@ARTICLE{Islam_ROC:2022,
  author={Islam, Riadul},
  journal={IEEE Canadian Journal of Electrical and Computer Engineering}, 
  title={{Early Stage DRC Prediction Using Ensemble Machine Learning Algorithms}}, 
  year={2022},
  volume={45},
  number={4},
  pages={354-364},
  doi={10.1109/ICJECE.2022.3200075}
}

@article{Fawcett:2006,
title = {{An introduction to ROC analysis}},
journal = {Pattern Recognition Letters},
volume = {27},
number = {8},
pages = {861-874},
year = {2006},
note = {ROC Analysis in Pattern Recognition},
issn = {0167-8655},
doi = {https://doi.org/10.1016/j.patrec.2005.10.010},
url = {https://www.sciencedirect.com/science/article/pii/S016786550500303X},
author = {Tom Fawcett},
}

@article{Hanley:1982,
  author = {Hanley, J. A. and McNeil, B. J.},
  title = {{The meaning and use of the area under a receiver operating characteristic (ROC) curve}},
  journal = {Radiology},
  volume = {143},
  number = {1},
  pages = {29--36},
  year = {1982},
  doi = {10.1148/radiology.143.1.7063747}
}

@INPROCEEDINGS{Islam_iscas:2014,
  author={Islam, Riadul and Guthaus, Matthew R.},
  booktitle={IEEE International Symposium on Circuits and Systems (ISCAS)}, 
  title={{Current-mode clock distribution}}, 
  year={2014},
  volume={},
  number={},
  pages={1203-1206},
  doi={10.1109/ISCAS.2014.6865357}
  }

@ARTICLE{Islam_cmcs:2017,
  author={Islam, Riadul and Guthaus, Matthew R.},
  journal={IEEE Transactions on Very Large Scale Integration (VLSI) Systems}, 
  title={CMCS: Current-Mode Clock Synthesis}, 
  year={2017},
  volume={25},
  number={3},
  pages={1054-1062},
  doi={10.1109/TVLSI.2016.2605580}}

@ARTICLE{Islam_HCDN:2019,
  author={Islam, Riadul and Guthaus, Matthew R.},
  journal={IEEE Transactions on Circuits and Systems I: Regular Papers}, 
  title={{HCDN: Hybrid-Mode Clock Distribution Networks}}, 
  year={2019},
  volume={66},
  number={1},
  pages={251-262},
  doi={10.1109/TCSI.2018.2866224}
  }

@misc{PROPHESEE:2024,
author = {PROPHESEE},
title = {{SONY-PROPHESEE IMX636 HD Event-Based Vision}},
howpublished = {\url{https://www.prophesee.ai/event-camera-evk4/}}
}

@article{ZhuVoxel,
  author       = {Alex Zihao Zhu and
                  Liangzhe Yuan and
                  Kenneth Chaney and
                  Kostas Daniilidis},
  title        = {{Unsupervised Event-based Learning of Optical Flow, Depth, and Egomotion}},
  journal      = {CoRR},
  volume       = {abs/1812.08156},
  year         = {2018},
  url          = {http://arxiv.org/abs/1812.08156},
  eprinttype    = {arXiv},
  eprint       = {1812.08156},
  bibsource    = {dblp computer science bibliography, https://dblp.org}
}

@inproceedings{RebecqVoxelFrame,
                title={{Real-time Visual-Inertial Odometry for Event Cameras using Keyframe-based Nonlinear Optimization}},
                author={Henri Rebecq, Timo Horstschaefer and Davide Scaramuzza},
                year={2017},
                month={September},
                pages={16.1-16.12},
                articleno={16},
                numpages={12},
                booktitle={Proceedings of the British Machine Vision Conference (BMVC)},
                publisher={BMVA Press},
                editor={Tae-Kyun Kim, Stefanos Zafeiriou, Gabriel Brostow and Krystian Mikolajczyk},
                doi={10.5244/C.31.16},
                isbn={1-901725-60-X},
                url={https://dx.doi.org/10.5244/C.31.16}
            }

@article{NTU:2019,
  title={Context Model for Pedestrian Intention Prediction Using Factored Latent-Dynamic Conditional Random Fields},
  author={Satyajit Neogi and Michael Hoy and K. Dang and Hang Yu and Justin Dauwels},
  journal={IEEE Transactions on Intelligent Transportation Systems},
  year={2019},
  volume={22},
  pages={6821-6832},
  url={https://api.semanticscholar.org/CorpusID:198968363}
}

@ARTICLE{WangSSIM2004,
  author={Zhou Wang and Bovik, A.C. and Sheikh, H.R. and Simoncelli, E.P.},
  journal={IEEE Transactions on Image Processing}, 
  title={{Image quality assessment: from error visibility to structural similarity}}, 
  year={2004},
  volume={13},
  number={4},
  pages={600-612},
  doi={10.1109/TIP.2003.819861}}

@article{Kim_tied:2024,
  title={{A tied-weight autoencoder for the linear dimensionality reduction of sample data}},
  author={Kim, Sunhee and Chu, Sang-Ho and Park, Yong-Jin and Lee, Chang-Yong},
  journal={Scientific Reports},
  volume={14},
  number={1},
  pages={26801},
  year={2024},
  publisher={Nature Publishing Group UK London}
}

@article{Zhou_denoising:2024,
  title={{Denoising-autoencoder-facilitated MEMS computational spectrometer with enhanced resolution on a silicon photonic chip}},
  author={Zhou, Jing and Zhang, Hui and Qiao, Qifeng and Chen, Heng and Huang, Qian and Wang, Hanxing and Ren, Qinghua and Wang, Nan and Ma, Yiming and Lee, Chengkuo},
  journal={Nature Communications},
  volume={15},
  number={1},
  pages={10260},
  year={2024},
  publisher={Nature Publishing Group UK London}
}

@misc{Ustek_anomaly:2024,
      title={{Deep Autoencoders for Unsupervised Anomaly Detection in Wildfire Prediction}}, 
      author={Irem Ustek and Miguel Arana-Catania and Alexander Farr and Ivan Petrunin},
      year={2024},
      eprint={2411.09844},
      archivePrefix={arXiv},
      primaryClass={cs.LG},
      url={https://arxiv.org/abs/2411.09844}, 
}

@article{rajput2024autoencoder,
  title={{An autoencoder-based deep learning model for solving the sparsity issues of Multi-Criteria Recommender System}},
  author={Rajput, Ishwari Singh and Tewari, Anand Shanker and Tiwari, Arvind Kumar},
  journal={Procedia Computer Science},
  volume={235},
  pages={414--425},
  year={2024},
  publisher={Elsevier}
}

@article{fettah2024convolutional,
  title={{Convolutional Autoencoder-Based medical image compression using a novel annotated medical X-ray imaging dataset}},
  author={Fettah, Amina and Menassel, Rafik and Gattal, Abdeljalil and Gattal, Abdelhak},
  journal={Biomedical Signal Processing and Control},
  volume={94},
  pages={106238},
  year={2024},
  publisher={Elsevier}
}

@INPROCEEDINGS{Islam_icrest:2023,
  author={Islam, Riadul and Majurski, Patrick and Kwon, Jun and Tummala, Sri Ranga Sai Krishna},
  booktitle={2023 3rd International Conference on Robotics, Electrical and Signal Processing Techniques (ICREST)}, 
  title={{Exploring High-Level Neural Networks Architectures for Efficient Spiking Neural Networks Implementation}}, 
  year={2023},
  volume={},
  number={},
  pages={212-216},
  doi={10.1109/ICREST57604.2023.10070080}
  }

@inproceedings{islam2025eaeventautoencoderhighspeed,
      title={EA: An Event Autoencoder for High-Speed Vision Sensing}, 
      author={Riadul Islam and Joey Mulé and Dhandeep Challagundla and Shahmir Rizvi and Sean Carson},
      year={2025},
      booktitle = {IEEE Computer Society Annual Symposium on VLSI (ISVLSI)},
}

@inproceedings{Kodukula_dvfs:2023,
author = {Kodukula, Venkatesh and Manetta, Mason and LiKamWa, Robert},
title = {Squint: A Framework for Dynamic Voltage Scaling of Image Sensors Towards Low Power IoT Vision},
year = {2023},
isbn = {9781450399906},
publisher = {Association for Computing Machinery},
address = {New York, NY, USA},
url = {https://doi.org/10.1145/3570361.3613303},
doi = {10.1145/3570361.3613303},
booktitle = {Proceedings of the 29th Annual International Conference on Mobile Computing and Networking},
articleno = {89},
numpages = {15},
location = {Madrid, Spain},
series = {ACM MobiCom}
}

@ARTICLE{Challagundla_tvlsi:2024,
  author={Challagundla, Dhandeep and Bezzam, Ignatius and Islam, Riadul},
  journal={IEEE Transactions on Very Large Scale Integration (VLSI) Systems}, 
  title={{ArXrCiM: Architectural Exploration of Application-Specific Resonant SRAM Compute-in-Memory}}, 
  year={2025},
  volume={33},
  number={1},
  pages={179-192},
  doi={10.1109/TVLSI.2024.3502359}
}

@phdthesis{islam2011high,
  title={{High-speed energy-efficient soft error tolerant flip-flops}},
  author={Islam, Riadul},
  year={2011},
  school={Concordia University}
}

@article{islam_dcmcs:2018,
  title={{DCMCS: Highly robust low-power differential current-mode clocking and synthesis}},
  author={Islam, Riadul and Fahmy, Hany A and Lin, Ping Y and Guthaus, Matthew R},
  journal={IEEE Transactions on Very Large Scale Integration (VLSI) Systems},
  volume={26},
  number={10},
  pages={2108--2117},
  year={2018},
  publisher={IEEE}
}

@ARTICLE{Islam_cjece:2019,
  author={Islam, Riadul},
  journal={Canadian Journal of Electrical and Computer Engineering}, 
  title={{Low-Power Highly Reliable SET-Induced Dual-Node Upset-Hardened Latch and Flip-Flop}}, 
  year={2019},
  volume={42},
  number={2},
  pages={93-101},
  doi={10.1109/CJECE.2019.2895047}
  }

@article{khwa2025mixed,
  title={{A mixed-precision memristor and SRAM compute-in-memory AI processor}},
  author={Khwa, Win-San and Wen, Tai-Hao and Hsu, Hung-Hsi and Huang, Wei-Hsing and Chang, Yu-Chen and Chiu, Ting-Chien and Ke, Zhao-En and Chin, Yu-Hsiang and Wen, Hua-Jin and Hsu, Wei-Ting and others},
  journal={Nature},
  volume={639},
  number={8055},
  pages={617--623},
  year={2025},
  publisher={Nature Publishing Group UK London}
}

@ARTICLE{Tossoun:2025,
  author={Tossoun, Bassem and Xiao, Xian and Cheung, Stanley and Yuan, Yuan and Peng, Yiwei and Srinivasan, Sudharsanan and Giamougiannis, George and Huang, Zhihong and Singaraju, Prerana and London, Yanir and Hejda, Matěj and Sundararajan, Sri Priya and Hu, Yingtao and Gong, Zheng and Baek, Jongseo and Descos, Antoine and Kapusta, Morten and Böhm, Fabian and Van Vaerenbergh, Thomas and Fiorentino, Marco and Kurczveil, Geza and Liang, Di and Beausoleil, Raymond G.},
  journal={IEEE Journal of Selected Topics in Quantum Electronics}, 
  title={{Large-Scale Integrated Photonic Device Platform for Energy-Efficient AI/ML Accelerators}}, 
  year={2025},
  volume={31},
  number={3: AI/ML Integrated Opto-electronics},
  pages={1-26},
  doi={10.1109/JSTQE.2025.3527904}
  }

@inproceedings{xi2026cnn,
  title={{CNN-Assisted Low-Power Clock Tree Synthesis for 3D ICs}},
  author={Xi, Chenbo and Zhou, Jindong and Zhou, Pingqiang},
  booktitle={2026 31st Asia and South Pacific Design Automation Conference (ASP-DAC)},
  pages={1421--1427},
  year={2026},
  organization={IEEE}
}

@article{rajendra2025ultra,
  title={{Ultra-Low-Power Dynamic-Bias Comparators With Self-Clocked Latch in 65-nm CMOS}},
  author={Rajendra, Anojh Kumaran and Bindra, Harijot Singh and Nauta, Bram},
  journal={IEEE Journal of Solid-State Circuits},
  year={2025},
  publisher={IEEE}
}

@inproceedings{vimala2025study,
  title={{A Study \& Analysis of Low Power Phase Locked Loop Design}},
  author={Vimala, G and VincyLloyd, F},
  booktitle={2025 IEEE 7th International Conference on Computing, Communication and Automation (ICCCA)},
  pages={1--4},
  year={2025},
  organization={IEEE}
}

@ARTICLE{Islam_neg:2021,
  author={Islam, Riadul},
  journal={IEEE Transactions on Emerging Topics in Computing}, 
  title={Negative Capacitance Clock Distribution}, 
  year={2021},
  volume={9},
  number={1},
  pages={547-553},
  doi={10.1109/TETC.2018.2872000}
  }

@INPROCEEDINGS{Islam_asicon:2011,
  author={Islam, Riadul and Esmaeili, S.E. and Islam, Thouhidul},
  booktitle={2011 9th IEEE International Conference on ASIC}, 
  title={{A high performance clock precharge SEU hardened flip-flop}}, 
  year={2011},
  volume={},
  number={},
  pages={574-577},
  doi={10.1109/ASICON.2011.6157270}}

@article{ambrogio2023analog,
  title={{An analog-AI chip for energy-efficient speech recognition and transcription}},
  author={Ambrogio, Stefano and Narayanan, Pritish and Okazaki, Atsuya and Fasoli, Andrea and Mackin, Charles and Hosokawa, Kohji and Nomura, Akiyo and Yasuda, Takeo and Chen, An and Friz, A and others},
  journal={Nature},
  volume={620},
  number={7975},
  pages={768--775},
  year={2023},
  publisher={Nature Publishing Group UK London}
}

\begin{IEEEbiography}[{\includegraphics[width=1in,height=1.25in,clip,keepaspectratio]{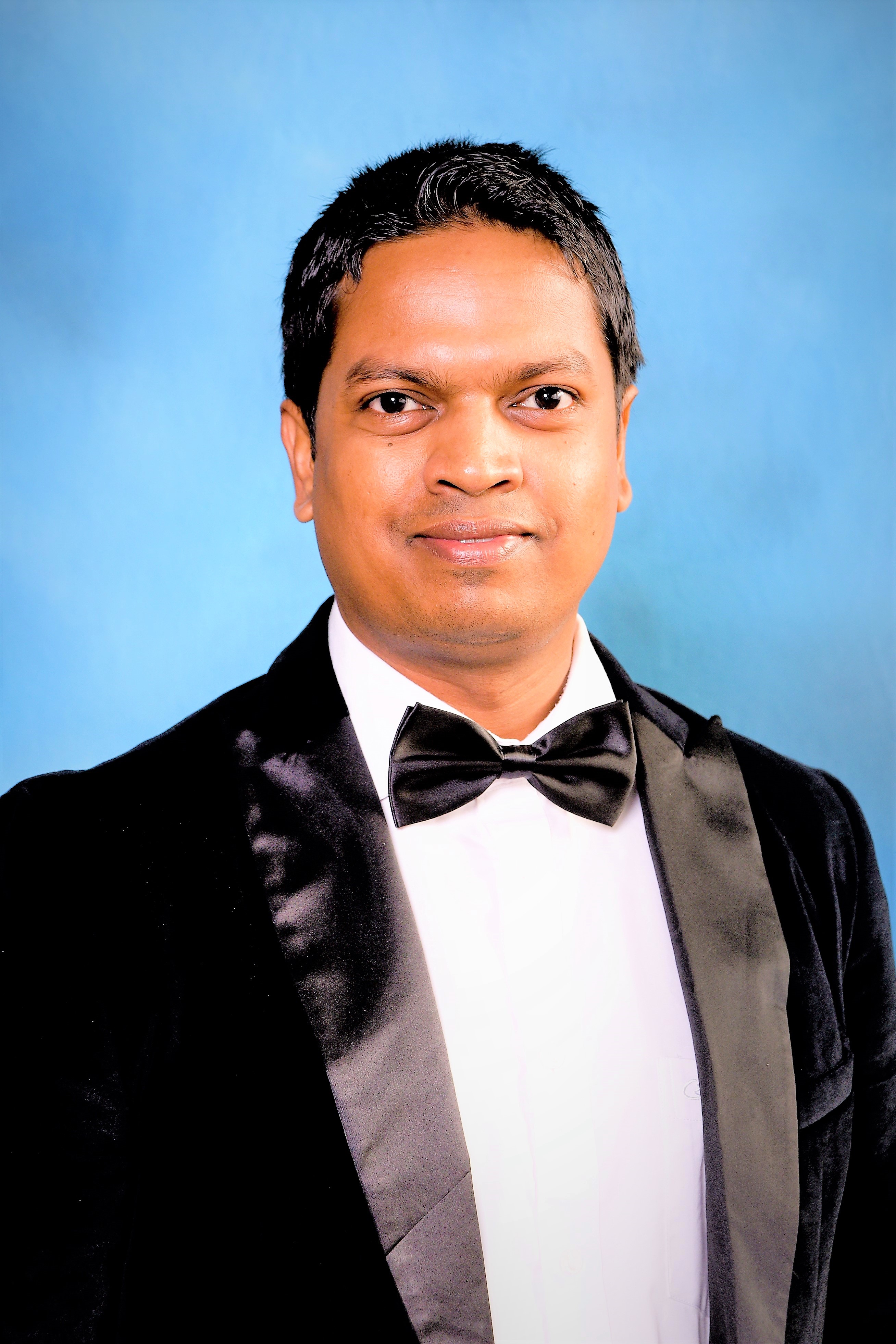}}]
{Riadul Islam}
	is currently a tenured associate professor in the
Department of Computer Science and Electrical Engineering at the University of Maryland, Baltimore County. 
In his Ph.D. dissertation work at UCSC, Riadul
designed the first current-pulsed flip-flop/register that resulted in the 
first-ever one-to-many current-mode clock distribution networks for
high-performance microprocessors. From 2017 to 2019, he was an Assistant
Professor with the University of Michigan, Dearborn MI, USA. 
He is a senior member of the IEEE, member of the ACM, IEEE Circuits and Systems (CAS) society, the VLSI Systems and Applications Technical 
Committee (VSA-TC) of the IEEE-CAS, and IEEE Solid-State Circuits (SSC) Society. 
He holds two US patent and several IEEE/ACM/MDPI/Springer Nature journal and conference publications. 
His  current  research
interests include  digital, analog, and mixed-signal CMOS ICs/SOCs for a 
variety of applications; verification and testing techniques for analog, 
digital and mixed-signal ICs; hardware security; CAN network; CAD tools for design and analysis of
microprocessors and FPGAs; automobile electronics; and biochips. 
He is an Associate Editor of Springer Circuits, Systems and Signal Processing (CSSP) Journal.
He was a Technical Program Committee (TPC) member of the IEEE/ACM International Conference on Computer-Aided Design (ICCAD 2022), 
ACM Great Lakes Symposium on VLSI (GLSVLSI 2020, GLSVLSI 2021, GLSVLSI 2022), 57th IEEE/ACM 
Design Automation Conference (DAC) 2020 LBR Session, IEEE Computer Society Annual Symposium on VLSI (ISVLSI) 2021,  and IEEE International Conference on Consumer Electronics (ICCE) 2021.
Riadul is the
recipient of a 2021 NSF ERI award, 2021 Maryland Industrial Partnerships (MIPS) award, and 2021 Maryland Innovation Initiative (MII) award.
\end{IEEEbiography}
\vspace{-1.25cm}
\begin{IEEEbiography}
[{\includegraphics[width=1in,height=1in,clip,keepaspectratio]{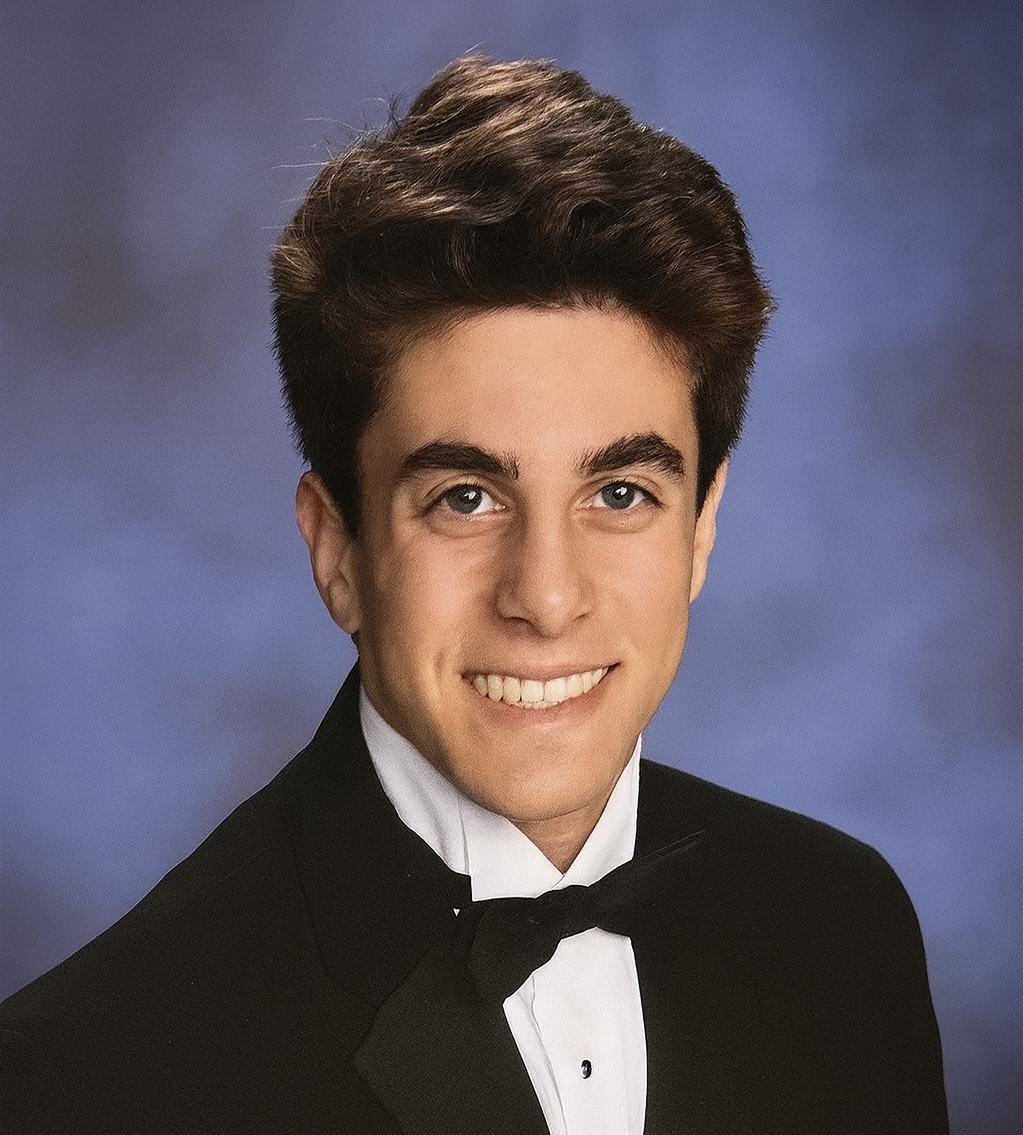}}] 
{Joey Mulé} (Student Member, IEEE) received his B.S. degree in Computer Science from the University of Maryland, Baltimore County (UMBC), MD, USA. He is currently pursuing the M.S. degree in Computer Science at UMBC, where he continues to explore research at the intersection of machine learning and artificial intelligence within the area of computer science and computer engineering. His academic and research interests include neural network architectural design, real-time object detection, adversarial defense systems, and low-power edge computing, with a particular focus on applications in event-based vision.
\end{IEEEbiography}
\vspace{-1.25cm}
\begin{IEEEbiography}
[{\includegraphics[width=1in,height=1in,clip,keepaspectratio]{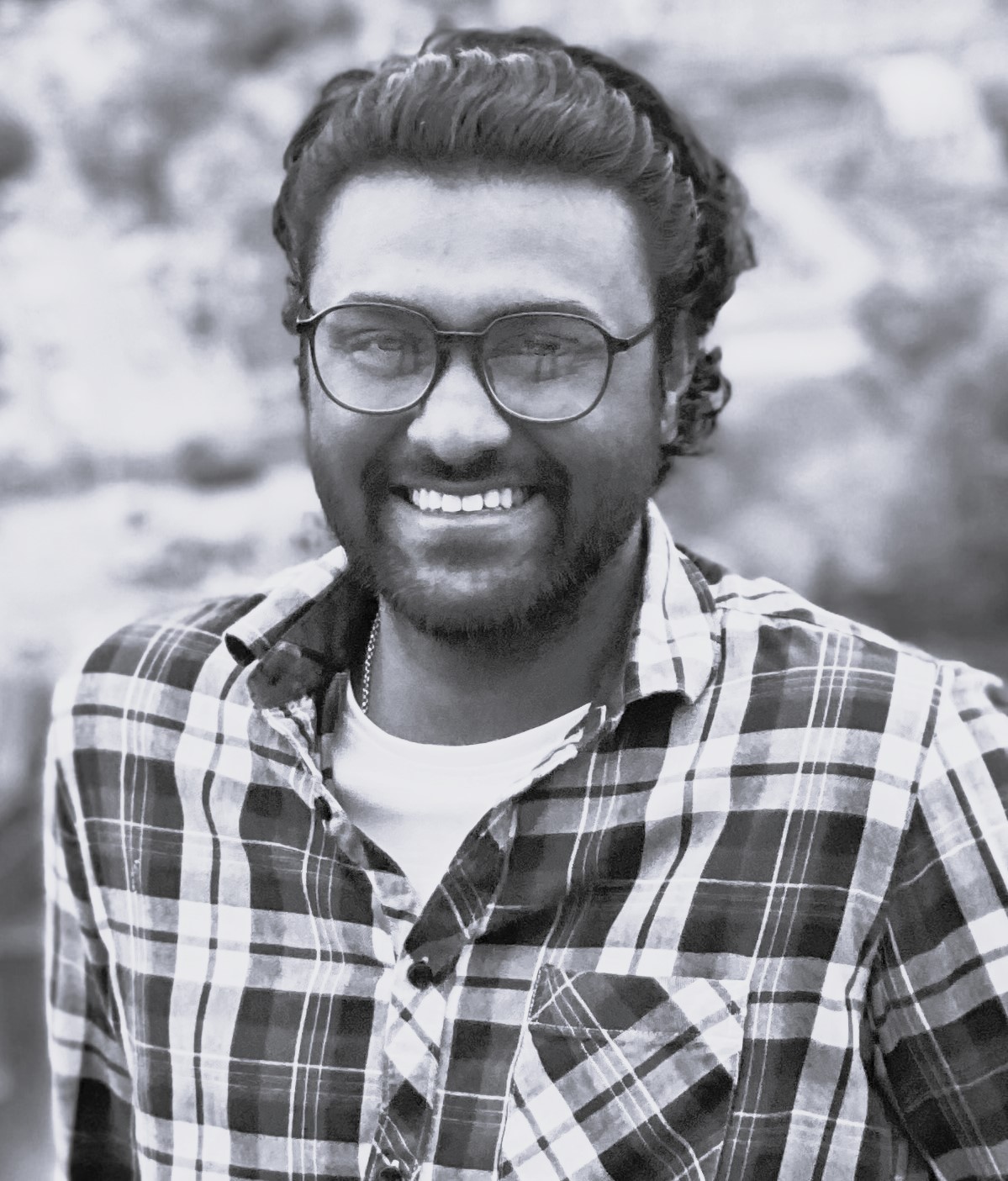}}] 
{Dhandeep Challagundla} (Student Member, IEEE) received his Ph.D. degree from The University of Maryland Baltimore County (UMBC), MD, USA. His research interests revolve around energy-efficient computing, Compute-in-Memories, SRAM design, low-power circuit design, Mixed-signal IC design, and EDA tools.
\end{IEEEbiography}
\vspace{-2.00cm}
\begin{IEEEbiography}
[{\includegraphics[width=1in,height=1in,clip,keepaspectratio]{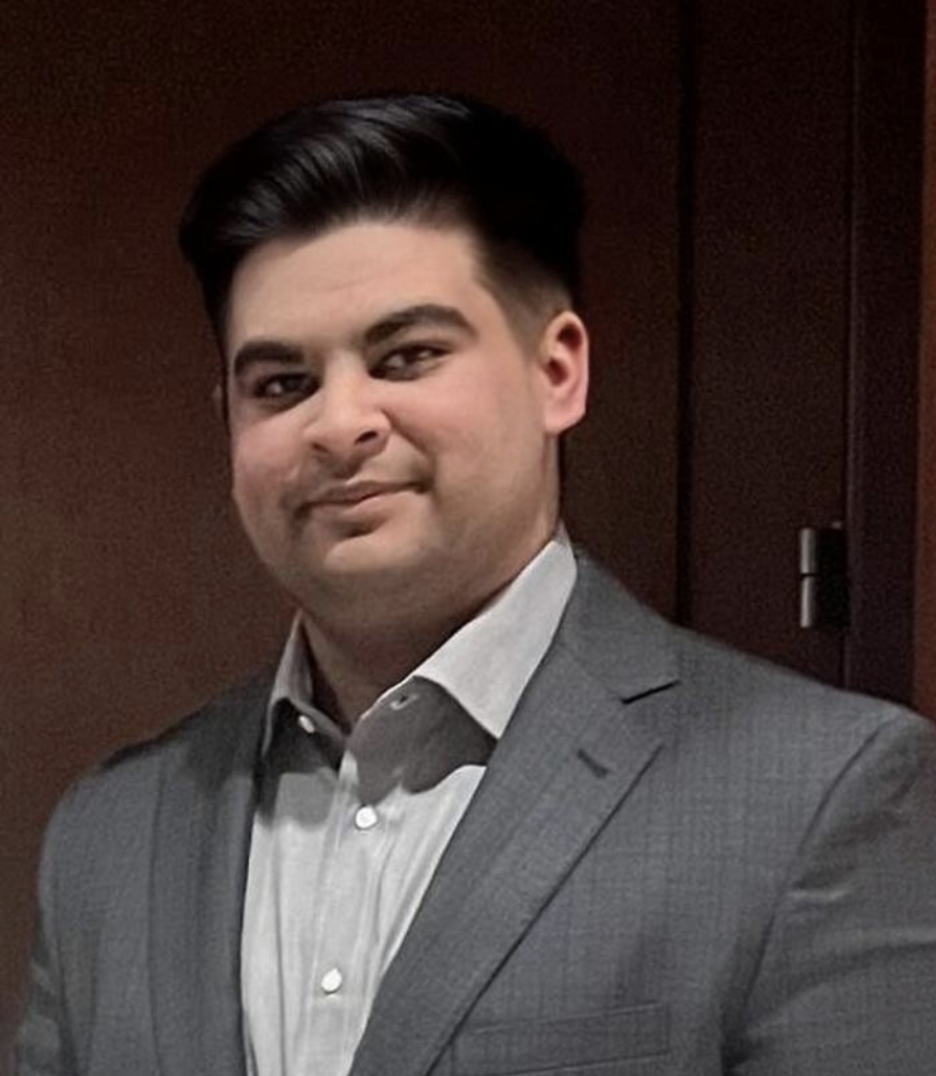}}] 
{Shahmir Rizvi} received his B.S. degree in Computer Engineering from the University of Maryland Baltimore County (UMBC), MD, USA, where he is currently pursuing his Ph.D. His research interests include VLSI design, embedded systems, and FPGA-based hardware acceleration.
\end{IEEEbiography}
\vspace{-2.00cm}
\begin{IEEEbiography}
[{\includegraphics[width=1in,height=1in,clip,keepaspectratio]{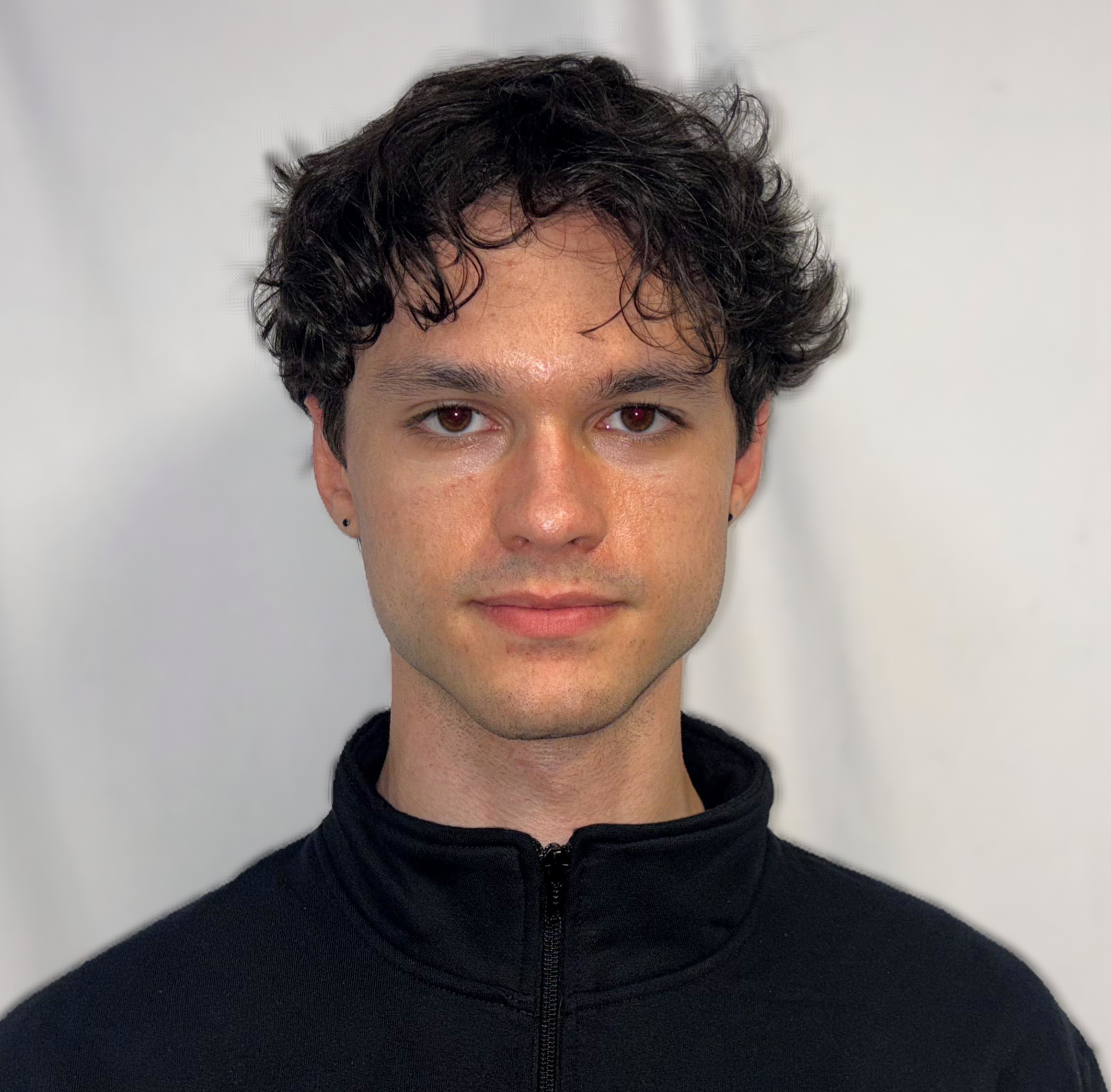}}] 
{Sean Carson} holds a B.S. degree in Computer Engineering from the University of Maryland, Baltimore County (UMBC), MD, USA. He currently researches at the UMBC VLSI-SOC Group, where his work focuses on event vision and neural networks. Outside of his research, he is employed as a Systems Engineer at Textron Systems, specializing in the development of electromagnetic environment simulators.
\end{IEEEbiography}

\vspace{-1.00cm}
\begin{IEEEbiography}
[{\includegraphics[width=1in,height=1in,clip,keepaspectratio]{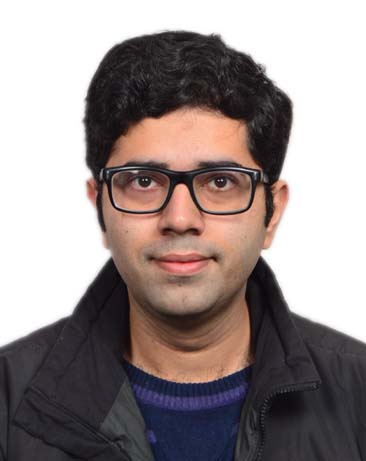}}] 
{Rachit Saini} completed his B.S. in Electrical Engineering from University of Illinois at Urbana-Champaign, Urbana-Champaign, IL, USA in 2014. He received his M.Tech. in 
VLSI Design from Thapar Institute of Engineering and Technology, Patiala, Punjab, India 
in 2018. Further, he received his M.S. in Computer Engineering from University of Maryland 
Baltimore County (UMBC), MD, USA in 2025. Currently he is pursuing his Ph.D. in Computer 
Engineering from UMBC. His current 
research interest includes VLSI design and security.
\end{IEEEbiography}




\end{document}